\documentclass[journal]{IEEEtran}

\usepackage{cite}

\ifCLASSINFOpdf
\usepackage[pdftex]{graphicx}

\else

\fi

\usepackage{amsmath}
\usepackage{amssymb}
\usepackage{amsfonts}
\usepackage{amsthm}
\usepackage{verbatim}
\usepackage{algorithmic}
\usepackage{algorithm}
\usepackage{subfigure}

\usepackage{array}

\begin{document}
%
\title{STSyn: Speeding Up Local SGD with Straggler-Tolerant Synchronization}

\author{Feng~Zhu,
        Jingjing~Zhang,~\IEEEmembership{Member,~IEEE}
        and~Xin~Wang,~\IEEEmembership{Fellow,~IEEE}
\thanks{The authors are with the Key Laboratory for Information Science of Electromagnetic Waves (MoE), Department of Communication Science and Engineering, School of Information Science and Technology, Fudan University, Shanghai 200433, China (e-mail: \{20210720072, jingjingzhang, xwang11\}@fudan.edu.cn).~(\textit{Corresponding author: Jingjing Zhang.})

This work was supported by the National Natural Science Foundation of China Grants No. 62101134 and No. 62071126, and the Innovation Program of Shanghai Municipal Science and Technology Commission Grant 20JC1416400.}}

\maketitle

\begin{abstract}
Synchronous local stochastic gradient descent (local SGD) suffers from some workers being idle and random delays due to slow and straggling workers, as it waits for the workers to complete the same amount of local updates. To address this issue, a novel local SGD strategy called STSyn is proposed in this paper. The key point is to wait for the $K$ fastest workers while keeping all the workers computing continually at each synchronization round, and making full use of any effective (completed) local update of each worker regardless of stragglers. To evaluate the performance of STSyn, an analysis of the average wall-clock time, average number of local updates, and average number of uploading workers per round is provided. The convergence of STSyn is also rigorously established even when the objective function is nonconvex for both homogeneous and heterogeneous data distributions. Experimental results highlight the superiority of STSyn over state-of-the-art schemes, thanks to its straggler-tolerant technique and the inclusion of additional effective local updates at each worker. Furthermore, the impact of system parameters is investigated. By waiting for faster workers and allowing heterogeneous synchronization with different numbers of local updates across workers, STSyn provides substantial improvements both in time and communication efficiency.
\end{abstract}

\begin{IEEEkeywords}
Local SGD, heterogeneous synchronization, straggler tolerance, distributed learning 
\end{IEEEkeywords}
\IEEEpeerreviewmaketitle


\section{Introduction}
\IEEEPARstart{A}{s} the size of datasets increases exponentially, it is no longer economic and feasible to leverage all the data to update the model parameters in large-scale machine learning, which magnifies the disadvantage of gradient descent (GD) methods in computational complexity. Instead, stochastic gradient descent (SGD) that uses just one or a mini-batch of samples is proved to converge faster and can achieve significantly reduced computation load \cite{bottou2010large,lian2017can,jiang2017collaborative}. Therefore, SGD has become the backbone of large-scale machine learning tasks.

To exploit parallelism and further accelerate the training process, the distributed implementation of worker nodes, i.e., distributed learning, has been gaining popularity\cite{li2014scaling,dean2012large,zinkevich2010parallelized}. The parameter server (PS) setting is one of the most common scenarios of SGD-based distributed learning frameworks, where in each iteration the PS aggregates the uploaded gradients/models from all the worker nodes to update the global parameter \cite{dean2012large,li2014scaling,lian2017can,sattler2019robust,cui2014exploiting,gupta2016model}. In such a setting, communication cost has become the bottleneck for the performance of the system due to the two-way communication and the larger number of parameters to be updated at each iteration \cite{goyal2017accurate,dube2019ai,dean2012large,smith2017federated,bottou2010large}. Accordingly, a novel scheme justified by the name local SGD is proposed to reduce the communication frequency between the PS and the workers by allowing each worker to perform local updates.

The idea of local SGD is first introduced in \cite{zinkevich2010parallelized} where each worker performs a certain number of local updates instead of one, and the PS aggregates the latest model of each worker as the final output. However, the scheme in \cite{zinkevich2010parallelized} has been proved to be no better than single-worker algorithms in the worst-case scenario due to the divergence of local optima across workers \cite{arjevani2015communication}. To address this issue, \cite{mcmahan2017communication} proposes the federated averaging (FedAvg) scheme, where each sampled worker runs multiple local updates before uploading the local model to the central server for aggregation and then the server re-sends the aggregated model back to the workers to repeat the above process. By increasing the communication frequency, FedAvg has been proved to be communication-efficient and enjoy remarkable performance under homogenous data distribution. The convergence analysis of the latter is widely explored in \cite{zhou2018convergence,stich2018local,wang2019cooperative,woodworth2018graph}. It should also be pointed out that FedAvg is proved to be able to achieve linear speedup concerning the number of workers in this setting. Furthermore, when data is heterogeneously distributed, the convergence of FedAvg is also proved by \cite{li2019convergence,khaled2019first,yu2019parallel,wang2019adaptive2,haddadpour2019convergence}. Local SGD has also been proved to achieve better time and communication performance than large mini-batch training with the size of the mini-batch being the same as the total batchsize used in a training round of local SGD \cite{lin2019don}.

\textbf{Adaptive Communication Strategies for Local SGD:} It has been proved by \cite{zhou2018convergence,yu2019parallel} that it is theoretically reasonable and advantageous to vary the communication frequency as the training proceeds. The authors of \cite{haddadpour2019local} present the convergence analysis of periodic averaging SGD (PASGD), which is FedAvg with no user sampling, and propose an adaptive scheme that gradually decreases the communication frequency between workers and the server to reduce total communication rounds. Meanwhile, \cite{wang2019adaptive} also proposes an adaptive strategy based on PASGD where the communication frequency is increased along the training process to achieve a decrease in the true convergence time. In addition, \cite{shen2021stl} proposes the STagewise Local SGD (STL-SGD) that increases the communication period along with decreasing learning rate to achieve further improvement.

Local SGD presents a possibility to reduce the communication cost of the system; yet, other than communication efficiency, the wall-clock time required for convergence is also an important issue. Due to the discrepancy among computing capabilities of machines, there is considerable variability in the computing time of different machines, and the machines that consume excessively more time per round than others are normally known as stragglers \cite{dean2013tail}. Numerous researches have been conducted in the development of straggler-tolerant techniques, which can be generally categorized into synchronous schemes and asynchronous ones. 

\textbf{Dealing with Stragglers with Synchronous SGD:} In synchronous parallel SGD, the PS always waits for all the workers to finish their computation. As a result, the whole process could be severely slowed down due to the existence of potential stragglers. One way to mitigate the effect of straggling is to exploit redundancy such as \cite{tandon2017gradient,li2017fundamental,wang2019erasurehead,zhang2020lagc,ozfatura2019gradient} that allow the PS to wait for only a subset of workers by utilizing storage and computation redundancy. Other than redundancy, \cite{harlap2016addressing} develops the FlexRR system that attempts to detect and avoid stragglers. Another well-known truth is that the convergence analysis of $M$-worker synchronous parallel SGD is the same as mini-batch SGD, only with the mini-batch size enlarged $M$ times. Inspired by this idea, \cite{gupta2016model} and \cite{dutta2018slow} propose the $K$-sync SGD where the PS only waits for the first $K$ workers to finish their computation. In addition, \cite{dutta2018slow} extends the $K$-sync SGD to $K$-batch-sync SGD in which the PS updates the parameter using the first $K$ mini-batches that finish, such that no worker is idle in each training round. 


In addition to the approaches mentioned above, there is a growing interest in allowing heterogeneous synchronization to deal with stragglers, where the number of local updates across workers varies in a synchronization round. Heterogeneous synchronization has been studied in several works, including \cite{rizk2020dynamic,wang2020tackling,ruan2021towards}. In \cite{rizk2020dynamic}, dynamic communication patterns are explored, where workers are manually designated to perform different numbers of local updates under a convex objective global function. FedNova, proposed in \cite{wang2020tackling}, balances the heterogeneity of local updates across workers by imposing weights inversely proportional to the number of local updates performed by the workers over the local models. Meanwhile, \cite{ruan2021towards} investigates the impact of incomplete local updates under heterogeneous data distribution by comparing the performances of three schemes. In that work, workers are required to perform a given number of local updates, and by the time of synchronization, those who have finished the task are considered to have ``complete" local updates, while the rest are considered to have ``incomplete" local updates. Scheme A only utilizes the complete local updates of the workers; Scheme B aggregates both complete and incomplete updates, and Scheme C is the canonical FedAvg scheme. However, all of \cite{rizk2020dynamic}, \cite{wang2020tackling}, and \cite{ruan2021towards} do not address the issue of fast workers being idle.

\textbf{Dealing with Stragglers with Asynchronous SGD:} Apart from confronting stragglers in the synchronous schemes, asynchronous SGD (ASGD) is also proposed to address this issue \cite{agarwal2011distributed}. In ASGD, the PS updates the parameter immediately after any worker finishes its computation and uploads it to the server. Apparently, the time used per round of ASGD schemes is much less than that used in synchronous SGD. Nevertheless, since the gradient used to update the parameter may not be the one used to compute it, the trade-off between wall-clock time and the staleness of the gradient becomes the main issue in ASGD, for which a number of variants of ASGD have been proposed. In \cite{recht2011hogwild}, the authors develop the Hogwild! scheme which is a lock-free implementation of ASGD and proves its effectiveness both theoretically and experimentall; and \cite{pan2016cyclades} improves the performance of ASGD based on Hogwild! with no conflict in parallel execution. In \cite{dutta2018slow}, the authors review the $K$-async SGD and $K$-batch-async SGD first introduced in \cite{gupta2016model} and \cite{lian2015asynchronous} respectively, and propose the AdaSync scheme to achieve the best error-runtime trade-off. Another trend of research is to process the delayed gradients in ASGD. To this end, \cite{sra2016adadelay} proposes the AdaDelay algorithm that imposes a penalty on delayed gradients, and \cite{zheng2017asynchronous} introduces the DC-ASGD scheme that compensates for the delayed gradients to make full use of all the gradients.

\subsection{Our Contributions}


This paper investigates the efficiency of straggler mitigation with heterogeneous synchronization both under homogenous and heterogeneous data distributions in a distributed PS-based architecture. The main contributions of our work are as follows:
\begin{itemize}
\item First, we propose a novel local SGD scheme that allows different numbers of local updates across workers while remaining robust to stragglers. Our approach builds on previous works that use heterogeneous synchronization, but we enhance it with straggler-tolerant techniques and the inclusion of all effective local updates from every worker to improve both time and communication efficiency.


\item Second, we provide an analysis of the average runtime, average number of uploading workers and average number of local updates per round, to justify the improvement in time and communication efficiency of our proposed scheme, named STSyn. We also rigorously establish the convergence of STSyn and prove that it can achieve a sublinear convergence rate, similar to other local SGD variants, but with a reduced convergence upper bound.

\item Finally, we present experimental results to corroborate the superiority of STSyn against state-of-the-art schemes for both homogeneous and heterogeneous data distributions in terms of wall-clock time and communication efficiency. We also study the influence of system hyper-parameters.
\end{itemize}



The rest of the paper is organized as follows. Section II describes the system model. The development of the proposed STSyn scheme is delineated in Section III. Section IV presents the analysis of STSyn. Numerical results are provided in Section V. Section VI concludes the work.

\textbf{Notations:} Boldface lowercase letters and calligraphic letters represent vectors and sets, respectively; $\mathbb{R}$ denotes the real number fields; $\mathbb{E}[\cdot]$ denotes the statistical expectation; $\bigcup$ denotes the union of sets; $\mathcal{A}\subseteq\mathcal{B}$ means that set $\mathcal{A}$ is a subset of set $\mathcal{B}$; $|\mathcal{A}|$ denotes the number of elements in set $\mathcal{A}$; $\nabla F$ represents the gradient of the function $F$;  $\|\mathbf{x}\|$ denotes the $\ell_2$-norm of vector $\mathbf{x}$; $\langle\mathbf{x},\mathbf{y}\rangle$ denotes the inner product of vectors $\mathbf{x}$ and $\mathbf{y}$.

\section{Problem Formulation}

\subsection{System Model}
We first introduce the general problem framework of distributed SGD with local updates in order to reduce communication overhead.
To implement the local-SGD-based training process, a parameter-server setting with a set of $M$ workers in set $\mathcal{M}\triangleq\left \{  1,...,M\right \}$ is considered. Each worker $m$ maintains a local dataset $\mathcal{D}_m$, which is equally allocated from the global training dataset, and we have  $\mathcal{D}=\bigcup_{m\in\mathcal{M}} \mathcal{D}_m$. Our objective is to minimize the following empirical risk function given as 
\begin{align}
F(\mathbf{w})=\frac{1}{M}\sum_{m=1}^{M} F_m(\mathbf{w}),\label{problem}
\end{align}
where variable $\mathbf{w}\in\mathbb{R}^d$ is the $d$-dimensional parameter to be optimized; and $F_m(\mathbf{w})$ is the local loss function of worker $m$ with $F_m(\mathbf{w})\triangleq\mathbb{E}[F_m(\mathbf{w};\xi_m)]$ where $\xi_i$ is drawn randomly from the local dataset $\mathcal{D}_m$. 

To elaborate on the communication and updating protocol, we define  $\mathbf{w}^j$ as the global parameter, and $\mathbf{w}_m^{j,0}$ as the local parameter that worker $m \in \mathcal{M}$ has available at training round $j$ prior to computation. 

\begin{figure}[hbt!]
\centering
\includegraphics[width=2in]{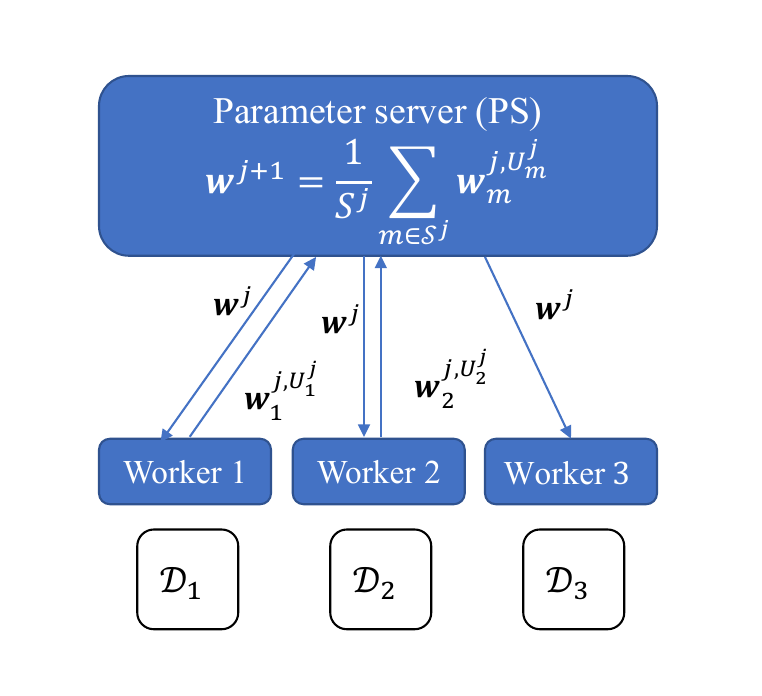}
\caption{Illustration of the PS setting with $M=3$, $\mathcal{M}^j=\{1,2,3\}$ and $\mathcal{S}^j=\{1,2\}$ at the $j$th round.}
\label{PS}
\end{figure}

In local SGD, at each synchronization round $j$, the PS first broadcasts the global model parameter $\mathbf{w}^{j}$ to a subset of workers $\mathcal{M}^j  \subseteq
\mathcal{M}$. Then by setting the local model as $\mathbf{w}_m^{j,0}=\mathbf{w}^j$, each worker $m$ in $\mathcal{M}^j$ starts to perform local updates. At the end of round $j$, each worker $m$ is assumed to complete $U_m^j\geq 0$ local updates, with each given as
\begin{equation}
    \mathbf{w}_{m}^{j,u+1}=\mathbf{w}_{m}^{j,u}-\frac{\alpha}{B} \sum_{b=1}^B \nabla F_m(\mathbf{w}_{m}^{j,u}; \xi_{m,b}^{j,u}), \label{update}
\end{equation}
for any $u=0,...,U_m^j-1$ if $U_m^j\geq 1$. Here $\alpha$ is the stepsize; the mini-batch $\{\xi_{m,b}^{j,u}\}_{b=1}^B$ of size $B$ is generated randomly for each local update from dataset $\mathcal{D}_m$; and $\nabla F_m(\mathbf{w}_{m}^{j,u}; \xi_{m,b}^{j,u})$ is the local gradient of loss function $F$ computed with local parameter $\mathbf{w}_{m}^{j,u}$ and data sample $\xi_{m,b}^{j,u}$. After the computation, the PS then selects a subset $\mathcal{S}^j \subseteq \mathcal{M}^j$ of $S^j$ workers to upload their local parameters,
yielding the global parameter $\mathbf{w}^{j+1}$ given as
\begin{align}
    \mathbf{w}^{j+1}=\frac{1}{S^j}\sum_{m\in\mathcal{S}^j} \mathbf{w}_{m}^{j,U_m^j}.\label{global_update}
\end{align}
The next round $j+1$ then starts and the training continues until a convergence criterion is satisfied, i.e., the trained model reaches a target test accuracy, with the total number of rounds defined as $J$. The whole process is illustrated in Fig.~\ref{PS}.

\subsection{Preliminaries} \label{background}

Based on the distributed PS architecture under consideration, our proposed scheme stems from a couple of local SGD variants, which we briefly review next.

\textbf{Federated Averaging (FedAvg).} One pioneering work on local update is the so-called FedAvg. More precisely, at each round, the PS first randomly selects a subset $\mathcal{M}^j\subseteq\mathcal{M}$ of workers to broadcast the global parameter to. Then each selected worker $m$ performs the same amount $U_m^j=U$ of local updates and sends the local parameter back to the PS, i.e., we have $\mathcal{S}^j=\mathcal{M}^j$. In particular, when all the workers participate in the computation in each round, i.e., with $\mathcal{S}^j=\mathcal{M}^j=\mathcal{M}$, we can arrive at periodic-averaging SGD (PASGD) \cite{haddadpour2019local}. 

\textbf{Adaptive Communication Strategy (AdaComm).} Building upon PASGD, \cite{wang2019adaptive} proposes an adaptive communication strategy called AdaComm in order to achieve a faster convergence speed in terms of wall-clock runtime. This is achieved by adaptively increasing the communication frequency between the PS and the workers. Note here that each worker still performs the same number of local updates in each training round.

\textbf{FedNova.} This is one of the pioneering works to investigate heterogeneous synchronization due to the discrepancy among the computing capabilities of machines \cite{wang2020tackling}, i.e., the numbers  $\{U_m^j\}_{m\in\mathcal{M}}$ of local updates across workers are not necessarily the same. FedNova aggregates the local models in a weighted manner to balance the heterogeneity of local updates, with the weight inversely proportional to the number of local updates the worker has completed.

\subsection{Performance Metrics}
In addition to training/learning accuracy, the time consumption and communication cost with the distributed learning scheme are also important performance metrics. Indeed, the training accuracy should be evaluated as a function of time and/or communication cost to fairly gauge the time/communication efficiency of a learning scheme. To this end, we specifically define wall-clock runtime and the communication cost as follows.

Firstly, the wall-clock runtime for each worker $m$ at the local iteration $u$ $(u< U_m^j)$ of round $j$ is defined as the time elapsed while computing the mini-batch gradient $\frac{1}{B}\sum_{b=1}^B \nabla F(\mathbf{w}_{m}^{j,u}; \xi_{m,b}^{j,u})$. It is denoted by $T_m^{j,u}$, and is assumed to conform to exponential distribution with mean $\mu$. We also assume that the random variable $T_m^{j,u}$ is independent across all workers, rounds and local iterations, which is a common assumption since workers are similar machines working independently \cite{tandon2017gradient,li2014scaling}. Therefore, the total time consumed at round $j$ can be written as 
\begin{align}
    T^j = \max_{m\in\mathcal{S}^j}\left\{\sum_{u=0}^{U_m^j-1}T_m^{j,u}\right\}. \label{wall-clock_time}
\end{align}

Secondly, notice that the global model parameter broadcast from the PS to workers and the local parameters sent from the workers to the PS, are of the same size. Hence, at each synchronization round $j$, we can simply define the communication cost $C^j$ as the sum of the number $|\mathcal{M}^j|$ of workers that download the global model from the PS, and the number $|\mathcal{S}^j|$ of workers that send the local update to the PS; i.e.,  
\begin{align}
    C^j = |\mathcal{M}^j| + |\mathcal{S}^j|.\label{communication}
\end{align}

These two performance metrics will be studied in the theoretical analysis in Section IV and will be used to evaluate the time/communication efficiency of the  schemes in the experimental results in Section V.

\begin{figure}[t!]
\centering
\includegraphics[width=3.6in]{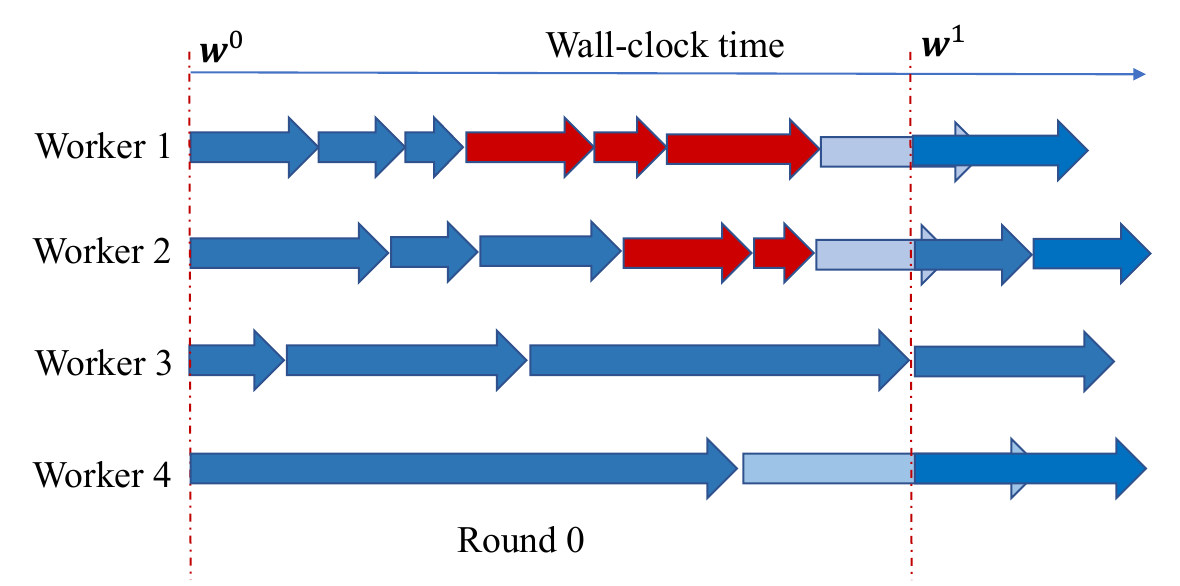}
\caption{Illustration of STSyn with $M=4$, $U=3$ and $K=3$. In round 0, worker 1, 2 and 3 are the fastest $K=3$ workers that have completed $U=3$ local updates; the red arrows represent the additional local updates performed by the fastest $K-1=2$ workers; and the arrows in light blue represent the straggling updates that are cancelled.}
\label{illus}
\end{figure}

\section{Local SGD with Heterogeneous Synchronization (STSyn)}

In this section, we introduce STSyn, a novel strategy that aims to accelerate local SGD both under i.i.d. and non-i.i.d. data with heterogeneous synchronization. The key points are two-fold. First, in each training round, no worker stops computing until the $K$th fastest one completes  a specific number $U$ of local updates. Hence, the faster the worker is, the more local updates it computes. Unlike typical local SGD methods where only the selected workers carry out the same number of local updates in each round, STSyn ensures that no worker remains idle, thus taking advantage of both straggler mitigation and training speedup.
Second, in the aggregation phase, any worker that completes at least one local update, i.e., with $U_m^j\geq 1$, sends its local parameter back to the PS. By increasing the number of workers that participate in the aggregation with each effective local update, STSyn can nicely balance the heterogeneity among workers and accelerate the training process. This also brings the benefit of communication efficiency in the sense that the reduction of synchronization rounds can compensate for the increase in communication cost at each round.

\begin{algorithm}[t]
\caption{STSyn}
\label{alg1}
\begin{algorithmic}
\STATE \textbf{Input:} number of workers $M$, constant $U,K$, stepsize $\alpha>0$, batch size $B$
\ENSURE 
\STATE Initiate $\mathbf{w}^0$
\FOR{$j=0,1,...J-1$}
\STATE Sends the global model $\mathbf{w}^j$ to all the workers in $\mathcal{M}$
\FOR{each worker $m\in \mathcal{M}$}
\STATE Executes WorkerUpdate$(m,\mathbf{w}^j)$
\ENDFOR
\IF{receives $K$ acknowledgements}
\STATE Stops the computation of the round and collects the local models $\{\mathbf{w}_{m}^{j,U_m^j}\}_{m\in\mathcal{S}^j}$ 
\ENDIF
\STATE Updates $\mathbf{w}^j$ with (\ref{global_update})
\ENDFOR

\vspace{15px}

\REQUIRE
\STATE Sets local model $\mathbf{w}_m^{j,0}=\mathbf{w}^j$ and $U_m^j=0$
\REPEAT
\STATE Updates its local model via (\ref{update}) and sets $U_m^j=U_m^j+1$
\IF {$U_m^j==U$}  
\STATE Sends an acknowledgment to the server
\ENDIF
\UNTIL{when noticed (by the server) the end of current round}
\IF{$U_m^j\geq 1$}
\STATE $\mathcal{S}^j\ni m$
\STATE Uploads its latest model $\mathbf{w}_{m}^{j,U_m^j}$ to the server 
\ENDIF
\end{algorithmic}
\end{algorithm}




Building on these ideas, we propose the novel
STSyn scheme as follows. In particular, at each training round $j$, the PS sends the current global model parameter $\mathbf{w}^j$ to all the workers in $\mathcal{M}$, i.e., we have $\mathcal{M}^j=\mathcal{M}$. Then each worker $m$ sets its local model parameter $\mathbf{w}_m^{j,0}=\mathbf{w}^j$ and begins its computation. After computing the first $U$ local updates, each worker acknowledges the completion but continues to update locally. Once the PS receives $K$ $ (K\leq M)$ acknowledgments, it terminates the computation of round $j$. However, for the rest $M-K$ straggling workers, they might have finished at least one local update, i.e., with $U_m^j\geq 1$, which can also be used for the global update. In other words, at round $j$, we might have a subset $\mathcal{S}^j\subseteq\mathcal{M}$ with $S^j\geq K$ workers that have completed at least one update. 

At the end of training round $j$, each worker in the subset $\mathcal{S}^j$ (instead of the set of the fastest $K$ workers) is required to upload its latest model parameter $\mathbf{w}_m^{j,U_m^j}$. Then, the PS updates the global parameter $\mathbf{w}^j$ through (\ref{global_update}). The next round $j + 1$ starts similarly and the training continues. An illustration of the STSyn scheme is shown in Fig.~\ref{illus}. Note that the cost and time elapsed of communicating the acknowledgment are assumed negligible here as they are typically much smaller than those with the local parameters. The complete procedure of STSyn is summarized in Algorithm 1.


\textit{Merits of STSyn:} 
The compound of the additional local updates from both the fastest $K$ workers and the rest $M-K$ stragglers brings multiplicative benefits; consequently, STSyn can outperform its local SGD counterparts. Intuitively, compared to FedAvg, with the choice $U_m^j=U$ and $\mathcal{M}^j=\mathcal{M}$, STSyn collects more local updates in a training round and also only has to wait for the fastest $K$ workers, thereby achieving better time and communication efficiency. 
Compared to FedNova, which relies on weighted heterogeneous synchronization, STSyn presents a distinct advantage without the need for such weights, as will be demonstrated in detail in Section V. In a nutshell, the straggler-tolerant techniques and the additional effective local updates together endow STSyn with the advantages both in wall-clock time and communication efficiency over state-of-the-art schemes.

\section{Performance Analysis}
In this section, we provide the analysis of the average runtime, average number of local updates and average number of uploading workers for each training round along with numerical illustrations. The benefits from additional effective local updates from the $K$ fastest workers and the rest $M-K$ stragglers are highlighted via comparison with PASGD. In addition, convergence of the proposed STSyn is rigorously established. 

\subsection{Average Wall-Clock Runtime}
As specified in Subsection II-C, we assume that the wall-clock times $\{T_m^{j,u}\}$ are i.i.d. exponential random variables across workers, rounds and local iterations, i.e., we have $T_m^{j,u} \sim {\rm Exp}(\mu)$ \cite{tandon2017gradient,li2014scaling}. 

At each round $j$, to carry out $U_m^j=U$ local updates, the runtime $T_m^j$ of each worker $m$ is given as $ T_m^j:=\sum_{u=0}^{U-1}T_m^{j,u}$. It then readily follows that $T_m^j$ is an Erlang random variable with mean $U\mu$ and variance $U\mu^2$. Since the PS only waits for the $K$th fastest worker to finish $U$ local updates, the average wall-clock time $\bar{T}$ taken per round can be computed as
\begin{align}
    \bar{T} = \mathbb{E}[T_{K:M}^j]=\frac{\mu M !}{(K-1) !(M-K) !} I 
    \label{time_per_iteration}
\end{align}
where $T_{K:M}^j$ is the $K$th order statistic of variables $\{T_m^j\}_{m\in\mathcal{M}}$, as analyzed in \cite{nadarajah2008explicit}, and we have the integral
\begin{equation}
  I=\int_{0}^{\infty}x\{Q(x)\}^{K-1}\{1-Q(x)\}^{M-K} q(x)dx,
\end{equation}
with $q(x)$ and $Q(x)$  being the functions
\begin{align}
    q(x)=\frac{x^{U-1}\exp(-x/\mu)}{\mu^U(U-1)!}, \notag\\
    Q(x)=\frac{\int_0^x t^{U-1}\exp(-t)dt}{(U-1)!}. \notag
\end{align}


\subsection{Average Number of Local Updates and Uploading Workers}


\begin{figure}[t!]
\centering
\includegraphics[width=3.5in]{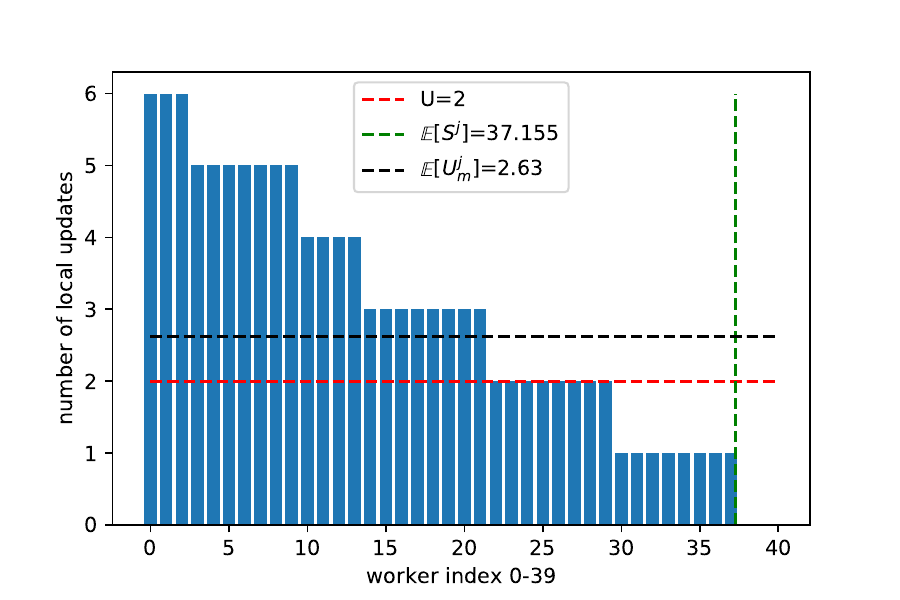}
\caption{The numbers of local updates across workers in a training round with $M=40,K=30$ and $U=2$.}
\label{examp}
\end{figure}

\textbf{Average Number of Local Updates per Round.} It is known that the average runtime per round is $ \mathbb{E}[T_{K:M}^j]$ and we have $T_m^{j,u} \sim {\rm Exp}(\mu)$. As a result, the average number $\bar{U}$ of local updates that each worker $m$ performs can be derived as
\begin{align}
    \bar{U}=\mathbb{E}[U_m^j]=\frac{\mathbb{E}[T_{K:M}^j]}{\mu}
    =\frac{M !}{(K-1) !(M-K) !} I.\label{average_number_lu}
\end{align}
It is easy to see that $\bar{U}$ increases as $K$ since $\mathbb{E}[T_{K:M}^j]$ does. On one hand, a larger $K$ indicates more local updates at each worker, which are beneficial for communication efficiency to some extent. On the other hand, the robustness against stragglers is decreased as we need to wait for more workers. Therefore, a trade-off between communication and time efficiency should be addressed, which is further explored through experiments in Section V.

\textbf{Average Number of Uploading Workers per Round.} Note that in STSyn, any worker that can finish at least one local update sends back its completed local parameter. To elaborate, for any integer $s\leq M$, we define as $T_{s:M}^{j,0}$ the $s$th order statistic of i.i.d. variables $\{T_{m}^{j,0}\}_{m=1}^{M}$, where $T_{m}^{j,0}$ is the runtime of worker $m$ consumed for the first local update. Then within runtime $T_{K:M}^j$ at round $j$, the number of selected workers $S^j$ can be given as
\begin{align}
   S^j= \max\left\{s\big|T_{s:M}^{j,0}\leq T_{K:M}^j\right\}. \label{number_of_S}
\end{align}
For exponential distribution, the average of $T_{s:M}^{j,0}$ can be approximated as \cite{dutta2018slow},
\begin{align}
    \mathbb{E}[T_{s:M}^{j,0}]=\sum_{i=M-s+1}^M\frac{\mu}{i}\approx \mu \log \frac{M}{M-s}. \label{exp_aver}
\end{align}
By using (\ref{exp_aver}) and (\ref{number_of_S}), 
$ \mathbb{E}[S^j] $ can be derived as
\begin{align}
    \mathbb{E}[S^j] \approx
    M - \exp\left(\log M- \frac{M !}{(K-1) !(M-K) !} I\right).\label{average_S}
\end{align}
It is easy to see that $\mathbb{E}[S^j]$ is also positively related to $K$ and $U$. That is to say, as the PS waits for more workers to finish more local iterations, a larger number of workers could complete at least one update and be selected to upload their local parameters at each round.   

Fig.~\ref{examp} shows an example of a certain training round of STSyn under the CIFAR-10 dataset using a CNN neural network with $M=40,K=30$, and $U=2$. It can be observed that almost half of the workers have completed more than $U=2$ local updates and 7 stragglers have one. These additional local updates are beneficial to the convergence speed, as will be illustrated in the sequel. Over 200 rounds, we also calculate the average number of local updates and that of workers selected to upload per round. As indicated in Fig.~\ref{examp}, the values are 2.63 and 37.155, both of them are consistent with the derived results $\mathbb{E}[U_m^j]=2.6352$ and $\mathbb{E}[S^j]= 37.132$ in (\ref{average_number_lu}) and (\ref{average_S}), respectively.

\begin{figure}[t!]
\centering
\includegraphics[width=3.5in]{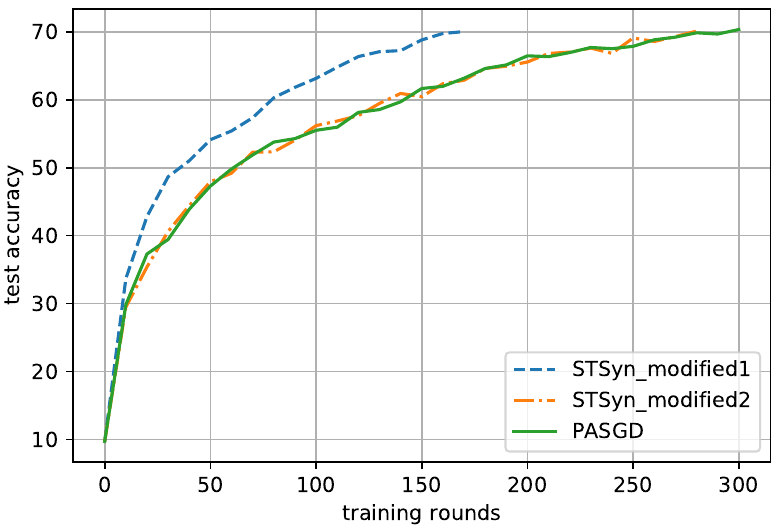}
\caption{Illustration of the impact of additional effective local updates on learning accuracy. In contrast to STSyn, STSyn\_modified1 is fixed with $S^j=K$; and STSyn\_modified2 is fixed with $U_m^j=U$.}
  \label{fig:eff}
\end{figure}

\textbf{Benefits from Additional Effective Local Updates.} By keeping all the workers computing until the end of each round, STSyn makes full use of any completed (effective) local update. This includes the additional ones performed both by the fastest $K$ workers and by the rest $M-K$ stragglers (see Fig.~\ref{examp}), as compared to the typical local SGD methods such as PASGD with $U_m^j=U$. 

The respective benefits of exploring both types are illustrated in Fig.~\ref{fig:eff}, where we trained a CNN on the CIFAR-10 dataset with $M=20, K=10$ and $U=5$. In particular, STSyn\_modified1 merely utilizes the additional local updates of the fastest workers, i.e., we have $U_m^j\geq U$ for the $K$ fastest workers and $S^j=K$; while STSyn\_modified2 only keeps the additional ones performed by the stragglers, i.e., we have $U_m^j=U$ for the $K$ fastest workers and $S^j\geq K$, which is equivalent to scheme B in \cite{ruan2021towards}. It can be observed that both of STSyn\_modified1 and STSyn\_modified2 require fewer training rounds than the baseline PASGD to reach the same test accuracy. In other words, any type of effective local update is favorable. It is also evidently seen that the ones of the fastest workers are indeed more useful. Meanwhile, the local updates from the stragglers are also contributive to balance the heterogeneity in synchronization and mitigate the client-drift effect \cite{zhao2018federated}. Furthermore, the joint effect of the two types is magnificent, which will be shown in Section V.

\subsection{Convergence Analysis}
We next rigorously establish the convergence of STSyn for both non-i.i.d. and i.i.d. data distributions. Our convergence analysis is based on the following standard assumptions, which are widely adopted in e.g., \cite{haddadpour2019convergence,stich2018local,li2019convergence,reddi2020adaptive}. 


\textbf{Assumption 1}  (Unbiasedness and Variance Boundedness) \textit{For fixed model parameter vector $\mathbf{w}$, the stochastic gradient $\nabla F_m(\mathbf{w}; \xi)$ is an unbiased estimator of the true gradient $\nabla F_m(\mathbf{w})$, i.e., we have}
\begin{equation}
    \mathbb{E}_{\xi}[\nabla F_m(\mathbf{w}; \xi)] = \nabla F_m(\mathbf{w}).\label{assumption2}
\end{equation}
 \textit{Moreover, there exists two scalars $\sigma_L>0$ and $\sigma_G>0$ such that}
\begin{align}
    \mathbb{E}_{\xi}\left[\left\|\nabla F_m(\mathbf{w}; \xi)-\nabla F_m(\mathbf{w})\right\|^2\right]\leq \sigma_L^2.\label{assumption31}\\
    \mathbb{E}\left[\left\|\nabla F_m(\mathbf{w})-\nabla F(\mathbf{w})\right\|^2\right]\leq \sigma_G^2.\label{assumption32}
\end{align}

\textbf{Assumption 2} (Smoothness and Lower Boundedness) \textit{The local loss function $F_m$ is $L$-smooth, i.e.,}
\begin{align}
    &\left\|\nabla F_m(\mathbf{w}_1)-\nabla F_m(\mathbf{w}_2)\right\| \leq L \left\|\mathbf{w}_1-\mathbf{w}_2\right\|,\label{lipschitz_gradient}\\
    &F_m(\mathbf{w}_1)\!-\! F_m(\mathbf{w}_2)\! \leq\! \left\langle\nabla F_m(\mathbf{w}_2), \mathbf{w}_1 \!- \!\mathbf{w}_2 \right\rangle \!+\! \frac{L}{2}\left\|\mathbf{w}_1 \!-\!\mathbf{w}_2\right\|^2,\label{lipschitz}
\end{align}
$\forall \mathbf{w}_1,\mathbf{w}_2\in\mathbb{R}^d.$ \textit{We also assume that the objective function $F$ is bounded below by a scalar $F^*$}.
Based on these assumptions, we can then derive an upper bound for the expectation of the average squared gradient norms $\frac{1}{J}\mathbb{E}\left[\sum_{j=0}^{J-1}\left\|\nabla F(\mathbf{w}^j)\right\|^2\right]$ of the objective function $F$, which is a common metric used for convergence analysis with general nonconvex objectives. 

\textbf{Theorem 1:} \textit{Let the constant stepsize $\alpha$ be chosen such that $\alpha\leq\min\big\{\frac{\bar{U}}{\sqrt{30}LU_{max}^2}, \frac{K^2}{L\bar{U}M^2}\big\}$. Under Assumptions 1 and 2, the expectation of the average squared norm of the gradients with heterogeneous data distribution can be upper bounded by}
\begin{align}
    &\frac{1}{J}\mathbb{E}\left[\sum_{j=0}^{J-1}\left\|\nabla F(\mathbf{w}^j)\right\|^2\right]\leq \frac{F(\mathbf{w}^J)-F^*}{c\alpha\bar{U}J}\notag\\
    &+\frac{1}{c}\left[\frac{\alpha U_{max}L}{2KB\bar{U}}\sigma_L^2 + \frac{5\alpha^2U_{max}^3L^2}{2\bar{U}^2}(\sigma_L^2+6U_{max}\sigma_G^2)\right]\label{theorem1},
\end{align}
where $c$ is a constant and we have bounded $U_m^j\leq U_{max}$, with $U_{max}:=\max\{U_m^j\}_{m\in\mathcal{M},0\leq j\leq J-1}$ , as the possibility that $U_m^j$ becomes unbounded is almost zero; we also used the expression  $\bar{U}=\mathbb{E}[U_m^j]$, as defined in Subsection IV-A.

\begin{proof}
The proof is presented in Appendix A. 
\end{proof}

\textbf{Corollary 1:} \textit{Under the conditions stated in Theorem 1, by selecting $\alpha=\frac{\sqrt{K}}{\sqrt{\bar{U}J}L}$, the convergence rate of STSyn for heterogeneous data distribution is} 
\begin{align}
\frac{1}{J}\mathbb{E}\left[\sum_{j=0}^{J-1}\left\|\nabla F(\mathbf{w}^j)\right\|^2\right]=\mathcal{O}\left(\frac{1}{\sqrt{KUJ}}+\frac{1}{J}\right).
\end{align}

Theorem 1 and Corollary 1 state that STSyn can achieve a sublinear convergence rate $\mathcal{O}\left(\frac{1}{\sqrt{KUJ}}+\frac{1}{J}\right)$, providing a theoretical guarantee for the performance of STSyn, which is consistent with the result of local SGD in heterogeneous data case. Since 
\cite{yang2020achieving} proves the fastest convergence rate of local SGD for nonconvex objective functions under heterogeneous data distribution, it would be a good baseline for comparison. Compared with the convergence upper bound in \cite{yang2020achieving} where each uploading worker performs the same number $U$ of local updates, here in (\ref{theorem1}) the role of $U$ is replaced with $\bar{U}$ and $U_{max}$, which are the average and upper bound of $U_m^j$, respectively. Since $\bar{U}$ is larger than $U$ as discussed in Section IV-B, the first term of the RHS of (\ref{theorem1}) is smaller, arriving at a faster convergence rate.




\textit{Remark:} 
The analysis of the i.i.d. data distribution follows a similar approach to the non-i.i.d. case. By setting $\sigma_G=0$ in Assumption 1, we can directly deduce the convergence results for the homogeneous distribution. Consequently, the convergence rate of STSyn under i.i.d. data can be expressed as $\mathcal{O}\left(\frac{1}{\sqrt{KUJ}}+\frac{1}{J}\right)$ by selecting $\alpha=\frac{\sqrt{K}}{\sqrt{\bar{U}J}L}$.


\section{Experiments}

In this section, we evaluate the effectiveness of the proposed STSyn. For comparison, we consider state-of-the-art schemes including FedNova \cite{wang2020tackling}, PASGD \cite{haddadpour2019local} and AdaComm  \cite{wang2019adaptive} mentioned in Subsection~\ref{background}. In addition, we explore the effect of the hyper-parameters $K$ and $U$ on the performance of STSyn.

\subsection{Experimental Setting and Metric Calculation}

\textbf{Experimental Setting.}
We consider training a three-layer CNN on the CIFAR-10 dataset with both homogenous and heterogeneous data distribution. To be specific, for the heterogenous case,
to distribute the training data samples equally among workers, we sort them by their labels. We set a total of $M=20$ workers for the training process. The mean of the exponential distribution of the computing time across workers is set to $\mu=10^{-4}$ sec. The stepsize $\alpha$ is 0.1 and the batch size $B$ is 100 for each scheme. 

\textbf{Metric Calculation.} We determine the per-round metrics, i.e., wall-clock time and communication cost, for FedNova, PASGD, AdaComm and the proposed STSyn, as follows. Note that, we generally take one synchronous communication from the workers to the PS as one round. 

\textit{Wall-clock time}. In FedNova, the number of local updates performed across workers conform to some distribution and the PS has to wait for all the workers. Hence, the wall-clock time consumed at round $j$ can be expressed as 
\begin{align}
    {T}_{FedNova}^j= Q_{M:M}^j,\label{time_fednova}
\end{align}
where $Q_m^j:=\sum_{u=0}^{U_m^j-1}T_m^{j,u}$ and $Q_{K:M}^j$ is the $K$th order statistic of $\{Q_m^j\}_{m\in\mathcal{M}}$.

For PASGD, each worker performs $U$ local updates per round and the PS waits for the slowest worker to finish. Therefore the per-round wall-clock time is 
\begin{align}
    {T}_{PASGD}^j= T_{M:M}^j.\label{time_pasgd}
\end{align}

For AdaComm, in each round $j$, each worker carries out $\tau^j$ local updates, with $\tau^j$ being decided adaptively and the PS also waits for all the workers. Then the per-round wall-clock time for AdaComm is
\begin{align}
    T_{AdaComm}^j=V_{M:M}^j, \label{time_adacomm}
\end{align}
where $V_m^j:=\sum_{u=0}^{\tau^j-1}T_m^{j,u}$ and $V_{K:M}^j$ is the $K$th order statistic of $\{V_m^j\}_{m\in\mathcal{M}}$.

Finally, as shown Section IV, the wall-clock time for STSyn at round $j$ is given as 
\begin{align}
    T_{STSyn}^j=T_{K:M}^j. \label{time_STSyn}
\end{align}

\textit{Communication cost}. For FedNova, in each round all the workers upload their local models. Therefore, the communication cost at round $j$ can be written as
\begin{align}
    C_{FedNova}^j=2M. \label{comm_fednova}
\end{align}

Similarly, the communication cost for PASGD is
\begin{align}
    C_{PASGD}^j=2M. \label{comm_pasgd}
\end{align}

For AdaComm, the per-round communication cost is
\begin{align}
    C_{AdaComm}^j=2M, \label{comm_adacomm}
\end{align}
since the PS has to wait for all the workers in $\mathcal{M}$.

The communication cost for STSyn at round $j$ is
\begin{align}
    C_{STSyn}^j=M+S^j, \label{comm_STSyn}
\end{align}
since in STSyn all the workers in $\mathcal{M}$ download the parameter $\mathbf{w}^j$ from the PS but only the ones in the subset $\mathcal{S}^j$ upload their local parameters.

\subsection{Experimental Results}
\begin{figure}[t!]
\centering
  \subfigure[i.i.d. distribution]{
		\includegraphics[width=3.5in]{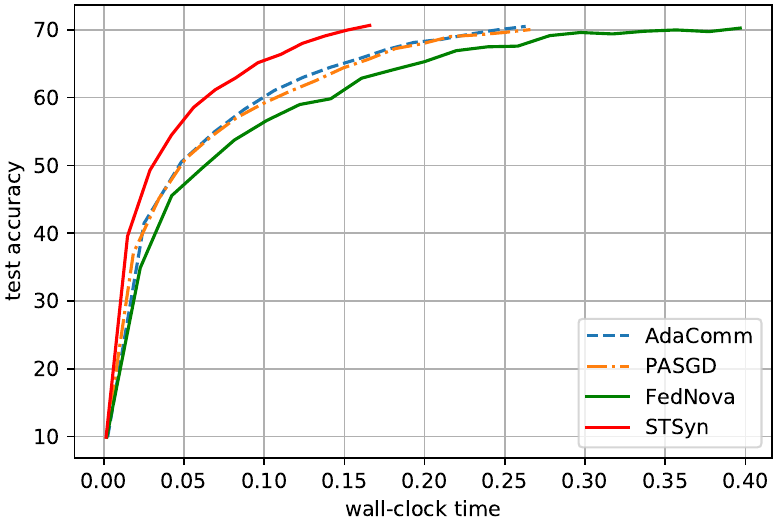}}

\subfigure[non-i.i.d. distribution]{
		\includegraphics[width=3.5in]{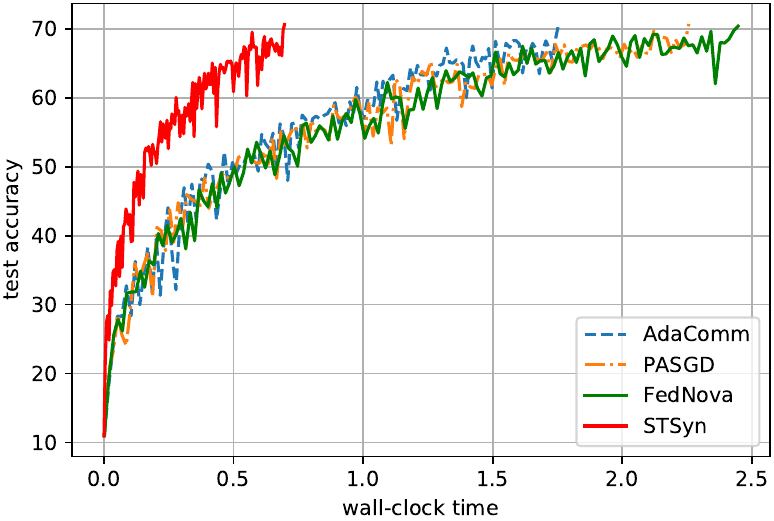}}

\caption{Comparison of test accuracy against wall-clock time for FedNova, AdaComm, PASGD and STSyn with a) i.i.d. distribution and b) non-i.i.d. distribution.}
\label{comp_time}
\end{figure}

\begin{figure}[t!]
\centering
\subfigure[i.i.d. distribution]{
		\includegraphics[width=3.5in]{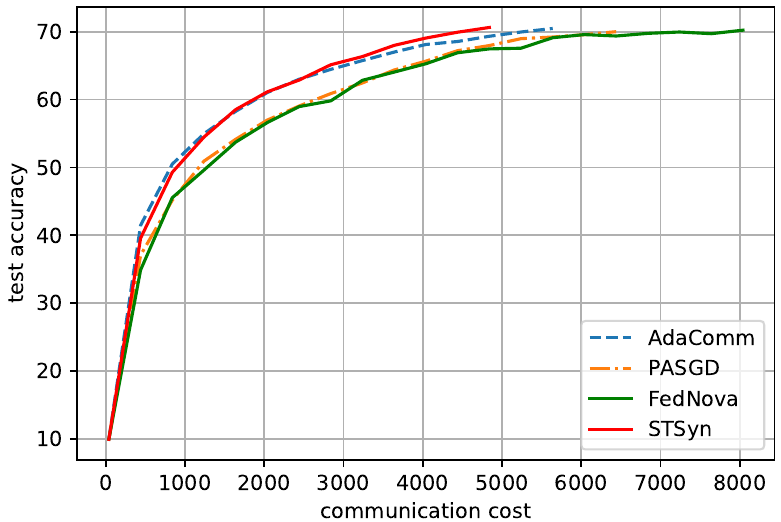}}
\subfigure[non-i.i.d. distribution]{
		\includegraphics[width=3.5in]{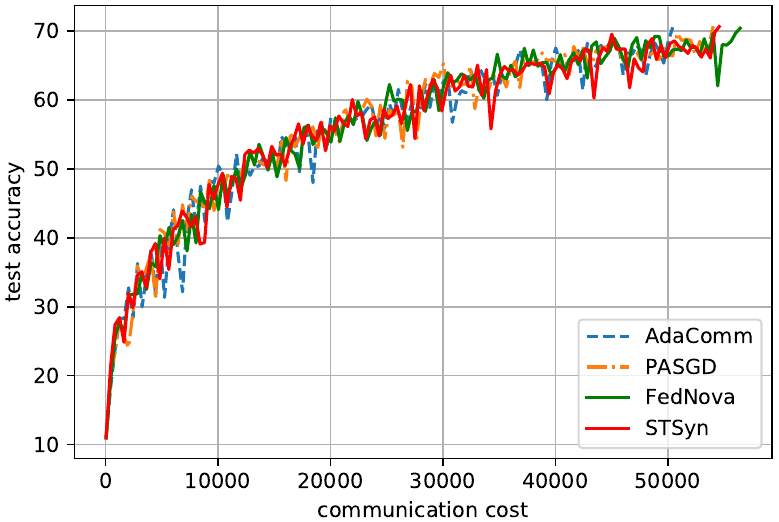}}
\caption{Comparison of test accuracy versus communication cost for FedNova, AdaComm, PASGD and STSyn with i.i.d. and non-i.i.d. distribution.}
\label{comp_comm}
\end{figure}

\textbf{STSyn Outperforms State-of-the-Art Schemes}. 
Fig.~\ref{comp_time} and \ref{comp_comm} compare the performance of FedNova, AdaComm, PASGD, and STSyn in terms of test accuracy for both the i.i.d. and non-i.i.d. cases. The local update numbers for FedNova conform to an exponential distribution with a mean of 10, while the initial number $U_{max}$ of local updates for AdaComm is set to be 10. PASGD uses a fixed number of 10 local updates, and STSyn is set with $K=5, U=10$ both cases. All schemes stop training once the accuracy reaches $70\%$. 


Fig.~\ref{comp_time} (a) and (b) provide a comprehensive comparison of the test accuracy of different schemes against wall-clock time under both homogeneous and heterogeneous data distributions. For both cases, it is clear that AdaComm performs better than PASGD by selecting the optimal communication frequency $U$ in different time intervals. In contrast, the weighted aggregation technique employed in FedNova proves to be ineffective in these scenarios, resulting in the poorest performance among the schemes considered. Out of all the schemes, STSyn stands out with the fastest training process, thanks to its incorporation of additional local updates and its robustness against stragglers, a key feature not present in the other three methods. It is worth noting that the curves corresponding to the i.i.d. distribution appear smoother compared to the other cases. This smoothness is attributed to the smaller variance inherent in the homogeneous data distribution.

Fig.~\ref{comp_comm} (a) and (b) present the performances of the communication efficiency exhibited by different schemes. In the i.i.d. case, STSyn consistently achieves the highest communication efficiency, thanks to the inclusion of additional effective local updates, particularly those with $U_m^j\leq U$. These additional updates effectively balance the synchronization process and contribute to the improved communication efficiency observed in STSyn. On the other hand, PASGD incurs a higher communication cost compared to both STSyn and AdaComm, primarily due to its limited utilization of local updates. FedNova performs the worst since it requires a larger number of rounds to achieve the target accuracy. 

Turning to the non-i.i.d. case, all four schemes achieve similar communication efficiency. This outcome is attributed to the client-drift phenomenon resulting from the data heterogeneity across workers, as discussed in prior studies \cite{zhao2018federated,karimireddy2020scaffold}. In other words, the communication efficiency does not consistently increase with the number of local updates.

In summary, the proposed STSyn algorithm can consistently outperform the other algorithms in terms of both time and communication efficiency across both data distribution cases.

\begin{figure}[t!]
\centering
\subfigure[i.i.d. distribution]{
		\includegraphics[width=3.5in]{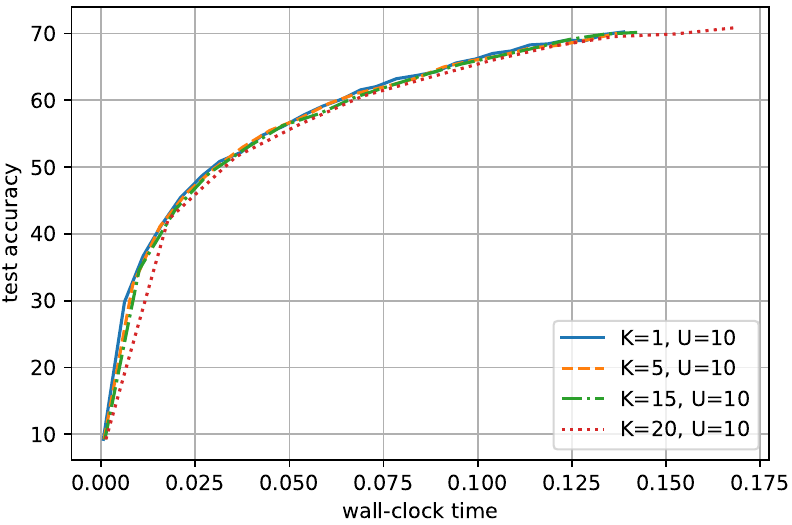}}
\subfigure[non-i.i.d. distribution]{
		\includegraphics[width=3.5in]{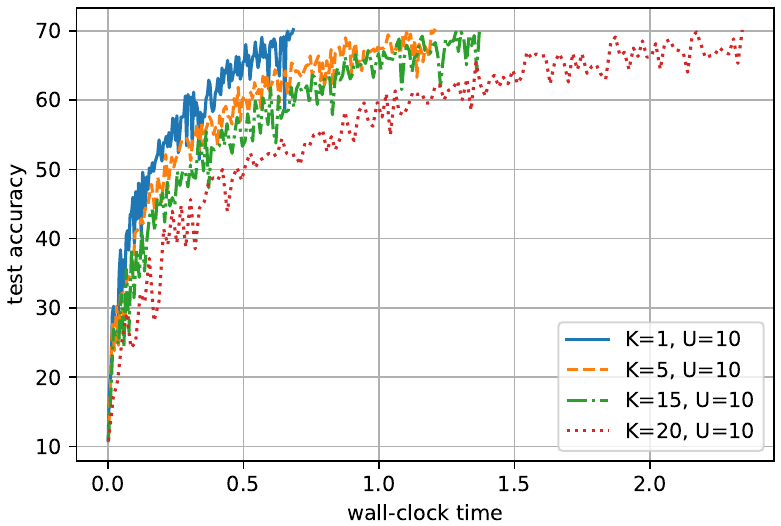}}
\caption{Influence of hyper-parameter $K$ of STSyn in terms of time-to-accuracy performance with i.i.d. and non-i.i.d. data distribution.}
\label{K_time}
\end{figure}

\begin{figure}[t!]
\centering
\subfigure[i.i.d. distribution]{
		\includegraphics[width=3.5in]{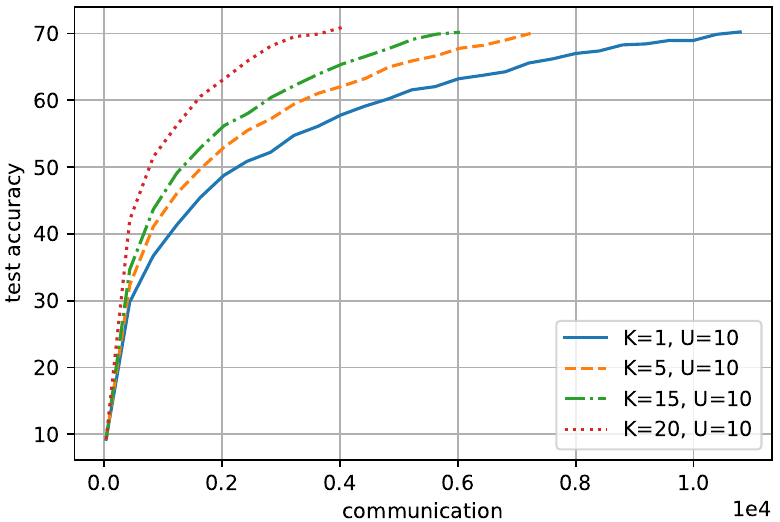}}
\subfigure[non-i.i.d. distribution]{
		\includegraphics[width=3.5in]{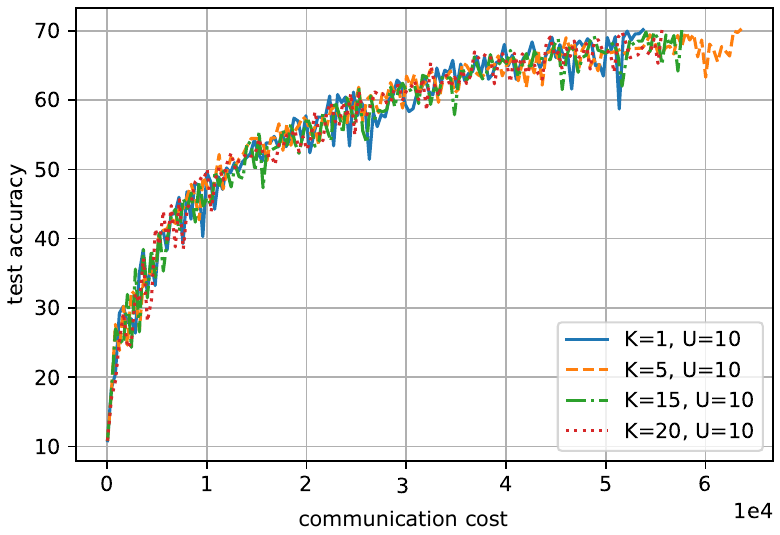}}
\caption{Influence of hyper-parameter $K$ of STSyn in terms of communication-to-accuracy performance with i.i.d. and non-i.i.d. data distribution.}
\label{K_comm}
\end{figure}

\begin{figure}[t!]
\centering
\subfigure[i.i.d. distribution]{
		\includegraphics[width=3.5in]{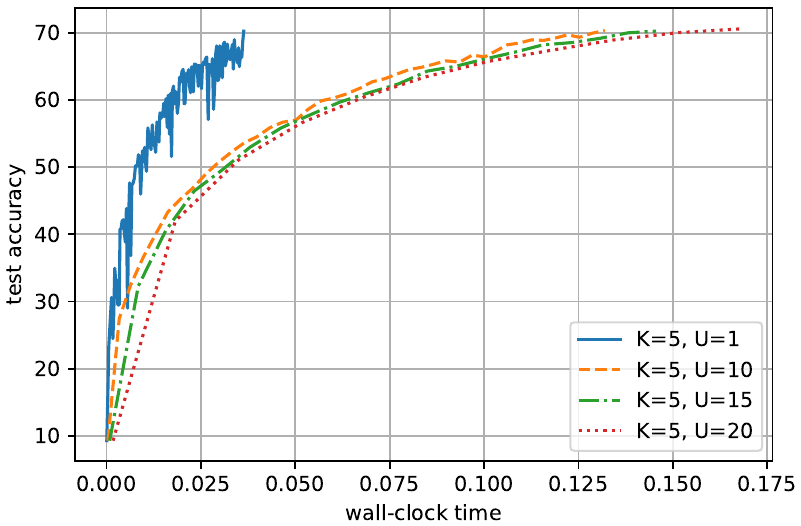}}
\subfigure[non-i.i.d. distribution]{
		\includegraphics[width=3.5in]{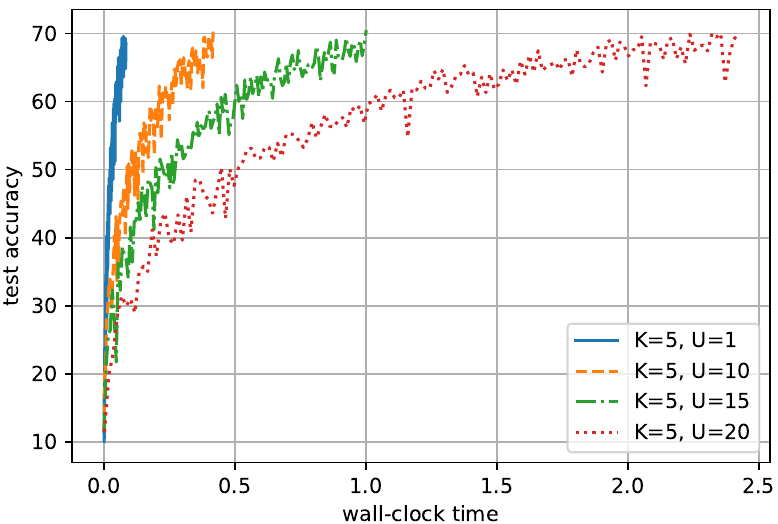}}
\caption{Influence of hyper-parameter $U$ of STSyn in terms of time-to-accuracy performance with i.i.d. and non-i.i.d. data distribution.}
\label{U_time}
\end{figure}

\textbf{Effect of Hyper-Parameter $K$}. 
By comparing STSyn schemes with varying values of $K$, we can observe the trade-off between wall-clock time and communication cost.


Fig.~\ref{K_time} (a) and (b) depict the performances of test accuracy against wall-clock time for different values of $K$ under both i.i.d. data and non-i.i.d. data distributions. In the case of i.i.d. data, the wall-clock time remains relatively stable when $K$ is small. As $K$ increases, the training process converges faster, leading to a reduction in the number of required rounds. This reduction offsets the increase in average round time caused by including more workers.
 However, it should be noted that as $K$ approaches the total number $M$, this compensation effect diminishes, resulting in an overall increase in wall-clock time. Conversely, under non-i.i.d. data distributions, the situation differs. As $K$ increases, the resulting speedup is unable to fully compensate for the reduction in the number of rounds due to the heterogeneity of the data. As a result, the wall-clock time performance degrades as $K$ increases. 


Fig.~\ref{K_comm} (a) and (b) demonstrate the communication-to-accuracy performance of STSyn for different values of $K$ under both i.i.d. and non-i.i.d. data distributions. In contrast to the observations regarding wall-clock time, it is obvious that a larger value of $K$, corresponding to a larger $S^j$, results in higher communication efficiency for i.i.d. data. This finding aligns with the analysis provided in Theorem 1, which explains that a larger $K$ accelerates convergence, and the reduction in the number of training rounds outweighs the increase in per-round communication cost. In the case of non-i.i.d. data, the variation in $K$ has minimal impact. This outcome primarily arises from the counterbalancing effects of the reduced number of training rounds and the increased per-round communication load.


In summary, when considering an i.i.d. data distribution, a larger value of $K$ is preferred as it leads to improved communication efficiency while maintaining a relatively high level of time efficiency. On the other hand, for a non-i.i.d. data distribution, a smaller value of $K$ is a better choice as it allows STSyn to achieve better time efficiency while still maintaining decent communication efficiency.

\textbf{Effect of Hyper-Parameter $U$}. As discussed in Section IV, there is a positive relationship between the value of $U$ and the number $S^j$ of workers selected per round. Consequently, increasing the value of $U$ results in the selection of a greater number of workers in each round.


Fig.~\ref{U_time} (a) and (b) showcase the time-to-accuracy performances of STSyn with various values of $U$ for both i.i.d. and non-i.i.d. data distributions. The figures reveal that the impact of $U$ differs from that of $K$. In both cases, the time-to-accuracy performance of STSyn consistently increases as $U$ decreases, indicating that the reduction in per-round wall-clock time through $U$ is a more influential factor in terms of time efficiency. These findings highlight that the choice of $U$ plays a crucial role in optimizing time efficiency in STSyn, regardless of the underlying data distribution.

\begin{figure}[t!]
\centering
\subfigure[i.i.d. distribution]{
		\includegraphics[width=3.5in]{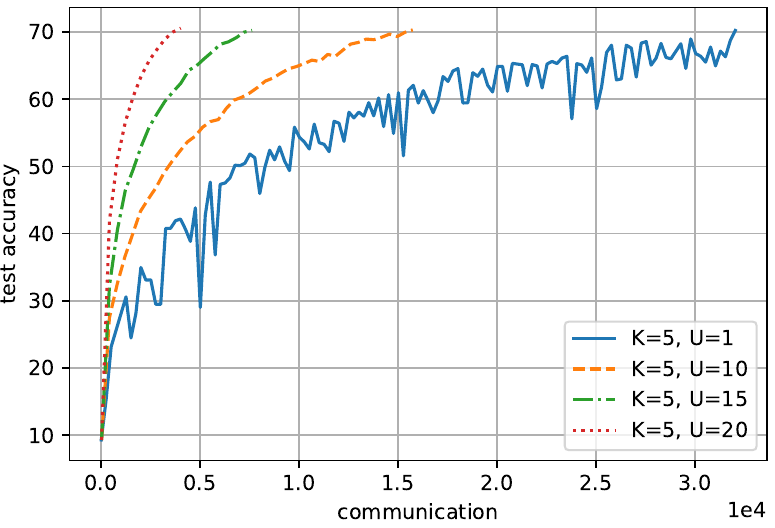}}
\subfigure[non-i.i.d. distribution]{
		\includegraphics[width=3.5in]{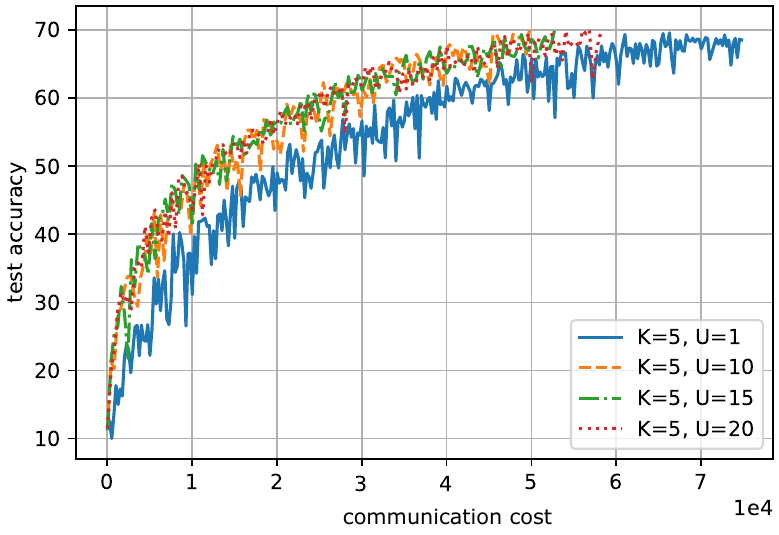}}
\caption{Influence of hyper-parameter $U$ of STSyn in terms of time-to-accuracy performance with i.i.d. and non-i.i.d. data distribution.}
\label{U_comm}
\end{figure}

Fig.~\ref{U_comm} (a) and (b) illustrate the communication-to-accuracy performances of STSyn. In the case of i.i.d. data, the communication efficiency increases as the value of $U$ becomes larger. This observation suggests that, for homogeneous data distribution, having more local updates per round leads to higher communication efficiency. However, in the case of non-i.i.d. data, increasing $U$ does not necessarily result in higher system communication efficiency. This is due to the client-drift effect, wherein more local updates cause the workers to diverge towards their own local optima.

Therefore, for homogeneous data distribution, selecting a moderate value of $U$ provides a preferable choice as it allows STSyn to strike a balance between time and communication efficiency. On the other hand, for heterogeneous data distribution, opting for a smaller value of $K$ is advisable in order to achieve better time efficiency and mitigate the client-drift effect.

In summary, the optimal selection of $U$ and $K$ depends on the specific characteristics of the data distribution. For homogeneous data, a moderate $U$ strikes the right balance, while for heterogeneous data, a smaller $K$ is preferred to enhance time efficiency and avoid the client-drift effect.

\section{Conclusion}
Building on a PS-based distributed learning framework for both i.i.d. and non-i.i.d. data distribution, we proposed STSyn, a novel local-SGD scheme with heterogeneous synchronization, which is both straggler-tolerant and communication-efficient. We derived the average wall-clock time, average number of local updates and average number of uploading workers for the proposed STSyn to substantiate the effectiveness of the additional local updates. The convergence of STSyn under homogenous and heterogeneous data distributions was also rigorously established. Extensive experimental results corroborated the superior time and communication efficiency of the proposed STSyn over the existing alternatives. The impacts of the hyper-parameters were also explored.


Our work leaves open a number of future research directions. First, it would be interesting to explore the potential advantages of heterogeneous synchronization for asynchronous implementations, where any worker can perform a number of local updates and send the result back to the PS. Second, generalization and analysis of the proposed STSyn scheme to the scenario where the wireless medium is considered, are worthy of further investigation. Lastly, integration of STSyn with the adaptive communication design 
might be also beneficial, and will be pursued in future work.

\bibliographystyle{IEEEtran}

\begin{thebibliography}{10}
\providecommand{\url}[1]{#1}
\csname url@samestyle\endcsname
\providecommand{\newblock}{\relax}
\providecommand{\bibinfo}[2]{#2}
\providecommand{\BIBentrySTDinterwordspacing}{\spaceskip=0pt\relax}
\providecommand{\BIBentryALTinterwordstretchfactor}{4}
\providecommand{\BIBentryALTinterwordspacing}{\spaceskip=\fontdimen2\font plus
\BIBentryALTinterwordstretchfactor\fontdimen3\font minus
  \fontdimen4\font\relax}
\providecommand{\BIBforeignlanguage}[2]{{%
\expandafter\ifx\csname l@#1\endcsname\relax
\typeout{** WARNING: IEEEtran.bst: No hyphenation pattern has been}%
\typeout{** loaded for the language `#1'. Using the pattern for}%
\typeout{** the default language instead.}%
\else
\language=\csname l@#1\endcsname
\fi
#2}}
\providecommand{\BIBdecl}{\relax}
\BIBdecl
\bibitem{bottou2010large}
L.~Bottou, ``Large-scale machine learning with stochastic gradient descent,''
  in \emph{Proc. COMPSTAT}, 2010, pp. 177--186.

\bibitem{lian2017can}
X.~Lian, C.~Zhang, H.~Zhang, C.~J. Hsieh, W.~Zhang, and J.~Liu, ``Can
  decentralized algorithms outperform centralized algorithms? {A} case study
  for decentralized parallel stochastic gradient descent,'' in \emph{Proc.
  Neural Inf. Process. Syst.}, vol.~30, 2017, pp. 5336--5346.

\bibitem{jiang2017collaborative}
Z.~Jiang, A.~Balu, C.~Hegde, and S.~Sarkar, ``Collaborative deep learning in
  fixed topology networks,'' in \emph{Proc. Neural Inf. Process. Syst.},
  vol.~30, 2017, pp. 5906--5916.

\bibitem{li2014scaling}
M.~Li, D.~G. Andersen, J.~W. Park, A.~J. Smola, and B.~Y. Su, ``Scaling
  distributed machine learning with the parameter server,'' in \emph{Proc.
  USENIX Symp. Oper. Syst. Des. Implement.}, 2014, pp. 583--598.

\bibitem{dean2012large}
J.~Dean, G.~S. Corrado, R.~Monga, C.~Kai, and A.~Y. Ng, ``Large scale
  distributed deep networks,'' in \emph{Proc. Neural Inf. Process. Syst.},
  vol.~25, 2012, pp. 1223--1231.

\bibitem{zinkevich2010parallelized}
M.~Zinkevich, M.~Weimer, A.~J. Smola, and L.~Li, ``Parallelized stochastic
  gradient descent,'' \emph{Proc. Neural Inf. Process. Syst.}, vol.~23, pp.
  2595--2603, 2010.

\bibitem{sattler2019robust}
F.~Sattler, S.~Wiedemann, K.-R. M{\"u}ller, and W.~Samek, ``Robust and
  communication-efficient federated learning from {Non}-iid data,'' \emph{IEEE
  Trans. Neural Netw. and Learn. Syst.}, vol.~31, no.~9, pp. 3400--3413, 2019.


\bibitem{cui2014exploiting}
H.~Cui, J.~Cipar, Q.~Ho, J.~Kim, S.~Lee, A.~Kumar, J.~Wei, D.~Wei, G.~R.
  Ganger, and P.~B. Gibbons, ``Exploiting bounded staleness to speed up big
  data analytics,'' in \emph{Proc. USENIX Annu. Tech. Conf.}, 2014, pp. 37--48.

\bibitem{gupta2016model}
S.~Gupta, W.~Zhang, and F.~Wang, ``Model accuracy and runtime tradeoff in
  distributed deep learning: A systematic study,'' in \emph{Proc. IEEE Int.
  Conf. Data Mining}, 2016, pp. 171--180.

\bibitem{goyal2017accurate}
P.~Goyal, P.~Dollár, R.~Girshick, P.~Noordhuis, L.~Wesolowski, A.~Kyrola,
  A.~Tulloch, Y.~Jia, and K.~He, ``Accurate, large minibatch {SGD}: Training
  imagenet in 1 hour,'' \emph{arXiv preprint arXiv:1706.02677}, 2017.

\bibitem{dube2019ai}
P.~Dube, T.~Suk, and C.~Wang, ``{AI} gauge: Runtime estimation for deep
  learning in the cloud,'' in \emph{Proc. IEEE Int. Symp. Comput. Architecture
  and High Perform. Comput.}, 2019, pp. 160--167.

\bibitem{smith2017federated}
V.~Smith, C.-K. Chiang, M.~Sanjabi, and A.~Talwalkar, ``Federated multi-task
  learning,'' in \emph{Proc. Neural Inf. Process. Syst.}, vol.~30, 2017, pp.
  4427--4437.

\bibitem{arjevani2015communication}
Y.~Arjevani and O.~Shamir, ``Communication complexity of distributed convex
  learning and optimization,'' in \emph{Proc. Neural Inf. Process. Syst.},
  vol.~28, 2015, pp. 1756--1764.

\bibitem{mcmahan2017communication}
H.~B. Mcmahan, E.~Moore, D.~Ramage, S.~Hampson, and B.~Arcas,
  ``Communication-efficient learning of deep networks from decentralized
  data,'' in \emph{Proc. Artif. Intell. and Statist.}, 2017, pp. 1273--1282.

\bibitem{zhou2018convergence}
F.~Zhou and G.~Cong, ``On the convergence properties of a {K}-step averaging
  stochastic gradient descent algorithm for nonconvex optimization,'' in
  \emph{Proc. Int. Joint Conf. Artif. Intell.}, 2018, pp. 3219--3227.
  
\bibitem{stich2018local}
S.~U. Stich, ``Local {SGD} converges fast and communicates little,'' in
  \emph{Proc. Proc. Int. Conf. Learn. Representations}, 2019.

\bibitem{wang2019cooperative}
J.~Wang and G.~Joshi, ``Cooperative {SGD}: A unified framework for the design
  and analysis of communication-efficient {SGD} algorithms,'' in \emph{Proc.
  ICML Workshop Coding Theory Mach. Learn.}, 2019.

\bibitem{woodworth2018graph}
B.~Woodworth, J.~Wang, B.~Mcmahan, and N.~Srebro, ``Graph oracle models, lower
  bounds, and gaps for parallel stochastic optimization,'' \emph{Proc. Neural
  Inf. Process. Syst.}, vol.~31, pp. 8505--8515, 2018.

\bibitem{li2019convergence}
X.~Li, K.~Huang, W.~Yang, S.~Wang, and Z.~Zhang, ``On the convergence of
  {FedAvg} on {Non-IID} data,'' in \emph{Proc. Int. Conf. Learn.
  Representations}, 2019.

\bibitem{khaled2019first}
A.~Khaled, K.~Mishchenko, and P.~Richt{\'a}rik, ``First analysis of local {GD}
  on heterogeneous data,'' \emph{arXiv preprint arXiv:1909.04715}, 2019.

\bibitem{yu2019parallel}
H.~Yu, S.~Yang, and S.~Zhu, ``Parallel restarted {SGD} with faster convergence
  and less communication: Demystifying why model averaging works for deep
  learning,'' in \emph{Proc. AAAI Conf. Artif. Intell.}, vol.~33, no.~01, 2019,
  pp. 5693--5700.

\bibitem{wang2019adaptive2}
S.~Wang, T.~Tuor, T.~Salonidis, K.~K. Leung, C.~Makaya, T.~He, and K.~Chan,
  ``Adaptive federated learning in resource constrained edge computing
  systems,'' \emph{IEEE J. Sel. Areas Commun.}, vol.~37, no.~6, pp. 1205--1221,
  2019.

\bibitem{haddadpour2019convergence}
F.~Haddadpour and M.~Mahdavi, ``On the convergence of local descent methods in
  federated learning,'' \emph{arXiv preprint arXiv:1910.14425}, 2019.

\bibitem{reddi2020adaptive}
S.~J. Reddi, Z.~Charles, M.~Zaheer, Z.~Garrett, K.~Rush, J.~Kone{\v{c}}n{\`y},
  S.~Kumar, and H.~B. McMahan, ``Adaptive federated optimization,'' in
  \emph{Proc. ICLR}, 2020.

\bibitem{lin2019don}
L.~Tao, S.~U. Stich, and M.~Jaggi, ``Don't use large mini-batches, use local
  {SGD},'' in \emph{Proc. Int. Conf. Learn. Representations}, 2019.



\bibitem{haddadpour2019local}
F.~Haddadpour, M.~M. Kamani, M.~Mahdavi, and V.~R. Cadambe, ``Local {SGD} with
  periodic averaging: Tighter analysis and adaptive synchronization,'' in
  \emph{Proc. Neural Inf. Process. Syst.}, vol.~32, 2019, pp. 11\,082--11\,094.

\bibitem{wang2019adaptive}
J.~Wang and G.~Joshi, ``Adaptive communication strategies to achieve the best
  error-runtime trade-off in local-update {SGD},'' in \emph{Proc. Mach. Learn.
  and Syst.}, vol.~1, 2019, pp. 212--229.

\bibitem{shen2021stl}
S.~Shen, Y.~Cheng, J.~Liu, and L.~Xu, ``{STL-SGD}: Speeding up local {SGD} with
  stagewise communication period,'' in \emph{Proc. AAAI Conf. Artif. Intell.},
  vol.~35, no.~11, 2021, pp. 9576--9584.

  

\bibitem{dean2013tail}
J.~Dean and L.~A. Barroso, ``The tail at scale,'' \emph{Commun. ACM}, vol.~56,
  no.~2, pp. 74--80, 2013.

\bibitem{rizk2020dynamic}
E.~Rizk, S.~Vlaski, and A.~H. Sayed, ``Dynamic federated learning,''\emph{arXiv preprint arXiv:2002.08782},
 2020.
 
\bibitem{wang2020tackling}
J.~Wang, Q.~Liu, H.~Liang, G.~Joshi, and H.~V. Poor, ``Tackling the objective
  inconsistency problem in heterogeneous federated optimization,'' in
  \emph{Proc. NeurIPS}, vol.~33, 2020, pp. 7611--7623.

\bibitem{ruan2021towards}
Y.~Ruan, X.~Zhang, S.-C. Liang, and C.~Joe-Wong, ``Towards flexible device
  participation in federated learning,'' in \emph{Proc. Int. Conf. Artif.
  Intell. and Statist.}, 2021, pp. 3403--3411.


\bibitem{tandon2017gradient}
R.~Tandon, L.~Qi, A.~G. Dimakis, and N.~Karampatziakis, ``Gradient coding:
  Avoiding stragglers in distributed learning,'' in \emph{Proc. Int. Conf.
  Mach. Learn.}, 2017, pp. 3368--3376.

\bibitem{li2017fundamental}
S.~Li, M.~A. Maddah-Ali, Q.~Yu, and A.~S. Avestimehr, ``A fundamental tradeoff
  between computation and communication in distributed computing,'' \emph{IEEE
  Trans. Inf. Theory}, vol.~64, no.~1, pp. 109--128, 2017.
  
\bibitem{wang2019erasurehead}
H.~Wang, Z.~Charles, and D.~Papailiopoulos, ``Erasurehead: Distributed gradient
  descent without delays using approximate gradient coding,'' \emph{arXiv
  preprint arXiv:1901.09671}, 2019.

\bibitem{zhang2020lagc}
J.~Zhang and O.~Simeone, ``Lagc: Lazily aggregated gradient coding for
  straggler-tolerant and communication-efficient distributed learning,''
  \emph{IEEE Trans. Neural Netw. and Learn. Syst.}, vol.~32, no.~3, pp.
  962--974, 2020.

\bibitem{ozfatura2019gradient}
E.~Ozfatura, D.~G{\"u}nd{\"u}z, and S.~Ulukus, ``Gradient coding with
  clustering and multi-message communication,'' in \emph{Proc. IEEE Data Sci.
  Workshop}, 2019, pp. 42--46.

\bibitem{harlap2016addressing}
A.~Harlap, H.~Cui, W.~Dai, J.~Wei, G.~R. Ganger, P.~B. Gibbons, G.~A. Gibson,
  and E.~P. Xing, ``Addressing the straggler problem for iterative convergent
  parallel {ML},'' in \emph{Proc. ACM Symp. Cloud Comput.}, 2016, pp. 98--111.



\bibitem{dutta2018slow}
S.~Dutta, J.~Wang, and G.~Joshi, ``Slow and stale gradients can win the race:
  Error-runtime trade-offs in distributed {SGD},'' in \emph{Proc. Int. Conf.
  Artif. Intell. and Statist.}, 2018, pp. 803--812.


\bibitem{agarwal2011distributed}
A.~Agarwal and J.~C. Duchi, ``Distributed delayed stochastic optimization,'' in
  \emph{Proc. Neural Inf. Process. Syst.}, vol.~24, 2011, pp. 873--881.

\bibitem{recht2011hogwild}
F.~Niu, B.~Recht, C.~Re, and S.~J. Wright, ``Hogwild!: A lock-free approach to
  parallelizing stochastic gradient descent,'' in \emph{Proc. Neural Inf.
  Process. Syst.}, vol.~24, 2011, pp. 693--701.

\bibitem{pan2016cyclades}
X.~Pan, M.~Lam, S.~Tu, D.~Papailiopoulos, C.~Zhang, M.~I. Jordan,
  K.~Ramchandran, C.~Re, and B.~Recht, ``Cyclades: Conflict-free asynchronous
  machine learning,'' in \emph{Proc. Neural Inf. Process. Syst.}, vol.~29,
  2016, pp. 2576--2584.

\bibitem{lian2015asynchronous}
X.~Lian, Y.~Huang, Y.~Li, and J.~Liu, ``Asynchronous parallel stochastic
  gradient for nonconvex optimization,'' in \emph{Proc. Neural Inf. Process.
  Syst.}, vol.~28, 2015, pp. 2737--2745.

\bibitem{sra2016adadelay}
S.~Sra, A.~W. Yu, M.~Li, and A.~J. Smola, ``Adadelay: Delay adaptive
  distributed stochastic optimization,'' in \emph{Proc. Artif. Intell. and
  Statist.}, 2016, pp. 957--965.

\bibitem{zheng2017asynchronous}
S.~Zheng, Q.~Meng, T.~Wang, W.~Chen, N.~Yu, Z.~M. Ma, and T.~Y. Liu,
  ``Asynchronous stochastic gradient descent with delay compensation,'' in
  \emph{Proc. Int. Conf. Mach. Learn.}, 2017, pp. 4120--4129.
\bibitem{zhao2018federated}
Y.~Zhao \emph{et~al.}, ``Federated learning with non-{IID} data,'' \emph{arXiv
  preprint arXiv:1806.00582}, 2018.

\bibitem{karimireddy2020scaffold}
S.~P. Karimireddy, S.~Kale, M.~Mohri, S.~Reddi, S.~Stich, and A.~T. Suresh,
  ``Scaffold: Stochastic controlled averaging for federated learning,'' in
  \emph{Proc. Int. Conf. Mach. Learn.}\hskip 1em plus 0.5em minus 0.4em\relax
  PMLR, 2020, pp. 5132--5143.

\bibitem{yang2020achieving}
H.~Yang, M.~Fang, and J.~Liu, ``Achieving linear speedup with partial worker
  participation in {non-IID} federated learning,'' in \emph{Proc. ICLR}, 2020.














\bibitem{nadarajah2008explicit}
S.~Nadarajah, ``Explicit expressions for moments of order statistics,''
  \emph{Statistics \& Probability Letters}, vol.~78, no.~2, pp. 196--205, 2008.











\end{thebibliography}

\onecolumn
\appendices
\section{Proof of Theorem 1}
\textit{Proof.} According to (\ref{lipschitz}) in Assumption 2, the one-step difference of the objective function can be written as
\begin{align}
& F(\mathbf{w}^{j+1}) - F(\mathbf{w}^{j}) \notag\\
    & \overset{(a1)}{\leq}\left\langle\nabla F(\mathbf{w}^j), \mathbf{w}^{j+1} - \mathbf{w}^{j} \right\rangle
 + \frac{L}{2}\left\|\mathbf{w}^{j+1}-\mathbf{w}^{j}\right\|^2 \notag\\
    &\overset{(a2)}{=}\left\langle \nabla F(\mathbf{w}^{j}), \mathbb{E}[g^j+\alpha \bar{U}\nabla F(\mathbf{w}^{j})-\alpha \bar{U}\nabla F(\mathbf{w}^{j})] \right\rangle + \frac{L}{2}\mathbb{E}[\left\|g^j\right\|^2]\notag\\
    &\overset{(a3)}{=} -\alpha \bar{U}\left\|\nabla F(\mathbf{w}^{j})\right\|^2 + {\left\langle \nabla F(\mathbf{w}^{j}), \mathbb{E}[{g}^j+\alpha \bar{U}\nabla F(\mathbf{w}^{j})] \right\rangle} + \frac{L}{2}{\mathbb{E}[\left\|g^j\right\|^2]},\notag\\
    &\overset{(a4)}{=} -\alpha \bar{U}\left\|\nabla F(\mathbf{w}^{j})\right\|^2 + \underbrace{\left\langle \nabla F(\mathbf{w}^{j}), \mathbb{E}[\bar{g}^j+\alpha \bar{U}\nabla F(\mathbf{w}^{j})] \right\rangle}_{T_1} + \frac{L}{2}\underbrace{\mathbb{E}[\left\|g^j\right\|^2]}_{T_2},\label{onestep}
\end{align}
where $(a2)$ holds because of the definition $g^j=\frac{1}{S^j}\sum_{m\in\mathcal{S}^j}g_m^j$ where $g_m^j=-\alpha\sum_{u=0}^{U_m^j-1}\frac{1}{B}\sum_{b=1}^B\nabla F_m(\mathbf{w}_{m}^{j,u}; z_{m,b}^{j,u})$; $(a3)$ is directly from $(a2)$ and $(a4)$ is due to the definition that $\bar{g}^j=\frac{1}{M}\sum_{m\in\mathcal{M}}g_m^j$ since the distribution of the computing capabilities of workers is homogenous.

Next, we will bound the expectation of the first and second term $T_1$ and $T_2$ in (\ref{onestep}), respectively. For the first term, we have
\begin{align}
    &\quad\left\langle \nabla F(\mathbf{w}^{j}), \mathbb{E}\left[\left(\bar{g}^j+\alpha \bar{U}\nabla F(\mathbf{w}^{j})\right)\right] \right\rangle\notag\\
    & \overset{(b1)}{=} \left\langle \nabla F(\mathbf{w}^{j}), \mathbb{E}\left[-\frac{1}{BM}\sum_{m=1}^M\sum_{u=0}^{U_m^j-1}\sum_{b=1}^B\alpha \nabla F_m(\mathbf{w}_{m}^{j,u};z_{m,b}^{j,u}) +\alpha \bar{U}\nabla F(\mathbf{w}^{j})\right] \right\rangle \notag\\
    & \overset{(b2)}{=} \left\langle \nabla F(\mathbf{w}^{j}), \mathbb{E}\left[-\frac{1}{M}\sum_{m=1}^M\sum_{u=0}^{U_m^j-1}\alpha \nabla F_m(\mathbf{w}_{m}^{j,u}) +\alpha \bar{U}\frac{1}{M}\sum_{m=1}^M\nabla F(\mathbf{w}^{j})\right] \right\rangle \notag\\
    & \overset{(b3)}{=} \left\langle\sqrt{\alpha \bar{U}} \nabla F(\mathbf{w}^{j}), -\frac{\sqrt{\alpha}}{\sqrt{\bar{U}}}\mathbb{E}\left[\sum_{m=1}^M \frac{1}{M} \sum_{u=0}^{U_m^j-1}\left(\nabla F_m(\mathbf{w}_{m}^{j,u}) - \nabla F(\mathbf{w}^j)\right)\right]\right\rangle\notag\\
    & \overset{(b4)}{\leq} \frac{\alpha \bar{U}}{2} \left\|\nabla F(\mathbf{w}^{j})\right\|^2 + \frac{\alpha}{2\bar{U}}\mathbb{E}\left[\left\|\frac{1}{M}\sum_{m=1}^M \sum_{u=0}^{U_m^j-1}\left(\nabla F_m(\mathbf{w}_{m}^{j,u}) - \nabla F(\mathbf{w}^j)\right)\right\|^2\right]-\frac{\alpha}{2\bar{U}M^2}\mathbb{E}\left\|\sum_{m=1}^M \sum_{u=0}^{U_m^j-1}\nabla F_m(\mathbf{w}_{m}^{j,u})\right\|^2\notag\\
    & \overset{(b5)}{\leq} \frac{\alpha \bar{U}}{2} \left\|\nabla F(\mathbf{w}^{j})\right\|^2 + \frac{\alpha}{2\bar{U}M}\sum_{m=1}^M U_m^j\sum_{u=0}^{U_m^j-1}\mathbb{E}\left[\left\|\nabla F_m(\mathbf{w}_{m}^{j,u}) - \nabla F(\mathbf{w}^j)\right\|^2\right]-\frac{\alpha}{2\bar{U}M^2}\mathbb{E}\left\|\sum_{m=1}^M \sum_{u=0}^{U_m^j-1}\nabla F_m(\mathbf{w}_{m}^{j,u})\right\|^2\notag\\
    & \overset{(b6)}{\leq} \frac{\alpha \bar{U}}{2} \left\|\nabla F(\mathbf{w}^{j})\right\|^2 + \frac{\alpha L^2}{2\bar{U}M}\sum_{m=1}^M U_m^j\sum_{u=0}^{U_m^j-1}\mathbb{E}\left[\left\|\mathbf{w}_{m}^{j,u}- \mathbf{w}^j\right\|^2\right]-\frac{\alpha }{2\bar{U}M^2}\mathbb{E}\left\|\sum_{m=1}^M \sum_{u=0}^{U_m^j-1}\nabla F_m(\mathbf{w}_{m}^{j,u})\right\|^2\notag\\
    & \overset{(b7)}{\leq} \left(\frac{\alpha \bar{U}}{2}+\frac{15\alpha^3L^2U_{max}^4}{\bar{U}}\right)\left\|\nabla F(\mathbf{w}^{j})\right\|^2+\frac{5\alpha^3U_{max}^3L^2}{2\bar{U}}(\sigma_L^2+6U_{max}\sigma_G^2)-\frac{\alpha }{2\bar{U}M^2}\mathbb{E}\left\|\sum_{m=1}^M \sum_{u=0}^{U_m^j-1}\nabla F_m(\mathbf{w}_{m}^{j,u})\right\|^2\notag\\
    &\overset{(b8)}{=} \left(\frac{\alpha \bar{U}}{2}+\frac{15\alpha^3L^2U_{max}^4}{\bar{U}}\right)\left\|\nabla F(\mathbf{w}^{j})\right\|^2+\frac{5\alpha^3U_{max}^3L^2}{2\bar{U}}(\sigma_L^2+6U_{max}\sigma_G^2)-\frac{\alpha }{2\bar{U}M^2}\mathbb{E}\left\|\sum_{m\in\mathcal{S}^j}\sum_{u=0}^{U_m^j-1}\nabla F_m(\mathbf{w}_{m}^{j,u})\right\|^2\notag\\\label{A1}
\end{align}
where $(b1)$ is from the definition of $\bar{g}^j$; $(b2)$ is due to the unbiasedness assumption; $(b3)$ is from direct computation; $(b4)$ uses the fact that $\langle\mathbf{x},\mathbf{y}\rangle=\frac{1}{2}[\|\mathbf{x}\|^2+\|\mathbf{y}\|^2-\|\mathbf{x}-\mathbf{y}\|^2]$; $(b5)$ comes from Cauchy-Schwartz inequality; $(b6)$ follows from the $L$-smoothness assumption; $(b7)$ is from \cite[Lemma 3]{reddi2020adaptive}, which proves that
\begin{align}
\mathbb{E}\left[\left\|\mathbf{w}_{m}^{j,u}-\mathbf{w}^j\right\|^2\right]\leq5U_{max}\eta^2(\sigma_L^2+6U_{max}\sigma_G^2)+30U_{max}^2\eta^2 \left\|\nabla F(\mathbf{w}^j)\right\|^2,\label{reddi}
\end{align}
under the condition that $\alpha\leq\frac{1}{8LU_{max}}$, where $\sigma_L$ and $\sigma_G$ are two constants defined in Assumption 1. Considering that $U_m^j$ varies among workers and $U_{max}$ is the maximal value, we can naturally arrive at the conclusion and $(b8)$ uses the fact that only the workers in subset $\mathcal{S}^j$ have completed effective local updates.

For the second term, we have
\begin{align}
    &\quad\mathbb{E}[\|g^j\|^2]\notag\\
    &\overset{(c1)}{=}\mathbb{E}\left[\left\|\frac{1}{S^j}\sum_{m\in\mathcal{S}^j}g_m^j\right\|^2\right]\notag\\
    &\overset{(c2)}{=}\mathbb{E}\left[\left\|\frac{1}{S^j}\sum_{m\in\mathcal{S}^j}\alpha\sum_{u=0}^{U_m^j-1}\frac{1}{B}\sum_{b=1}^B\nabla F_m(\mathbf{w}_{m}^{j,u}; z_{m,b}^{j,u})\right\|^2\right]\notag\\
    &\overset{(c3)}{\leq} \frac{\alpha^2}{(S^j)^2}\mathbb{E}\left[\left\|\sum_{m\in\mathcal{S}^j}\sum_{u=0}^{U_m^j-1}\frac{1}{B}\sum_{b=1}^B\nabla F_m(\mathbf{w}_{m}^{j,u}; z_{m,b}^{j,u})\right\|^2\right]\notag\\
    &\overset{(c4)}{=}\frac{\alpha^2}{(S^j)^2}\mathbb{E}\left[\left\|\sum_{m\in\mathcal{S}^j}\sum_{u=0}^{U_m^j-1}\left(\frac{1}{B}\sum_{b=1}^B\nabla F_m(\mathbf{w}_{m}^{j,u}; z_{m,b}^{j,u})-\nabla F_m(\mathbf{w}_{m}^{j,u})\right)\right\|^2\right]+\frac{\alpha^2}{(S^j)^2}\mathbb{E}\left[\left\|\sum_{m\in\mathcal{S}^j}\sum_{u=0}^{U_m^j-1}\nabla F_m(\mathbf{w}_{m}^{j,u})\right\|^2\right]\notag\\
    &\overset{(c5)}{\leq} \frac{\alpha^2}{(S^j)^2B^2}\sum_{m\in\mathcal{S}^j}\sum_{u=0}^{U_m^j-1}\sum_{b=1}^B\mathbb{E}\left[\left\|\nabla F_m(\mathbf{w}_{m}^{j,u}; z_{m,b}^{j,u})-\nabla F_m(\mathbf{w}_{m}^{j,u})\right\|^2\right]+\frac{\alpha^2}{(S^j)^2}\mathbb{E}\left[\left\|\sum_{m\in\mathcal{S}^j}\sum_{u=0}^{U_m^j-1}\nabla F_m(\mathbf{w}_{m}^{j,u})\right\|^2\right]\notag\\
    &\overset{(c6)}{\leq} \frac{\alpha^2 U_{max}}{S^j B}\sigma_L^2+\frac{\alpha^2}{(S^j)^2}\mathbb{E}\left[\left\|\sum_{m\in\mathcal{S}^j}\sum_{u=0}^{U_m^j-1}\nabla F_m(\mathbf{w}_{m}^{j,u})\right\|^2\right]\label{T2},
\end{align}
where $(c1)$ is due to the definition of $g^j$; $(c2)$ is due to the definition of $g_m^j$; $(c3)$ comes from direct computation; $(c4)$ uses the fact that $\mathbb{E}[\|x\|^2]=\mathbb{E}[\|x-\mathbb{E}[x]\|^2+\|\mathbb{E}[x]\|^2]$ and the unbiasedness assumption; $(c5)$ is due to the fact that $\mathbb{E}[\|x_1+...+x_n\|^2]=\mathbb{E}[\|x_1\|^2+...+\|x_n\|^2]$ if $x_i'$s are independent with zero mean and $(c6)$ uses Assumption 1.

Combining (\ref{T2}) and (\ref{A1}) into (\ref{onestep}), we readily have
\begin{align}
    & F(\mathbf{w}^{j+1}) - F(\mathbf{w}^{j})\notag\\
    &\overset{(d1)}{\leq}-\left(\frac{\alpha\bar{U}}{2}-\frac{15\alpha^3L^2U_{max}^4}{\bar{U}}\right)\left\|\nabla F(\mathbf{w}^{j})\right\|^2 +\frac{\alpha^2 U_{max}L}{2S^j B}\sigma_L^2 + \frac{5\alpha^3U_{max}^3L^2}{2\bar{U}}(\sigma_L^2+6U_{max}\sigma_G^2)\notag\\
    &+\left(\frac{\alpha^2L}{2(S^j)^2}-\frac{\alpha }{2\bar{U}M^2}\right)\mathbb{E}\left\|\sum_{m\in\mathcal{S}^j}\sum_{u=0}^{U_m^j-1}\nabla F_m(\mathbf{w}_{m}^{j,u})\right\|^2\notag\\
    &\overset{(d2)}{\leq} -\alpha\bar{U}\left(\frac{1}{2}-\frac{15\alpha^2L^2U_{max}^4}{\bar{U}^2}\right)\left\|\nabla F(\mathbf{w}^{j})\right\|^2+\frac{\alpha^2 U_{max}L}{2S^j B}\sigma_L^2 + \frac{5\alpha^3U_{max}^3L^2}{2\bar{U}}(\sigma_L^2+6U_{max}\sigma_G^2)\notag\\
    &\overset{(d3)}{\leq}-\alpha\bar{U}c\left\|\nabla F(\mathbf{w}^{j})\right\|^2+\frac{\alpha^2 U_{max}L}{2K B}\sigma_L^2 + \frac{5\alpha^3U_{max}^3L^2}{2\bar{U}}(\sigma_L^2+6U_{max}\sigma_G^2)\label{onestepnew}
\end{align}
where $(d1)$ is from direct computation; $(d2)$ follows from $\frac{\alpha^2L}{2(S^j)^2}-\frac{\alpha }{2\bar{U}M^2}\leq 0$ if $\alpha \leq \frac{K^2}{L\bar{U}M^2}$ and $(d3)$ holds because there exists a positive constant $c$ satisfying $\frac{1}{2}-\frac{15\alpha^2L^2U_{max}^4}{\bar{U}^2}>c$ if $\alpha\leq\frac{\bar{U}}{\sqrt{30}LU_{max}^2}$ and the fact that $S^j\geq K$.

Summing (\ref{onestepnew}) from $0$ to $J-1$ and rearranging the terms we have:
\begin{align}
    \frac{1}{J}\mathbb{E}\left[\sum_{j=0}^{J-1}\left\|\nabla F(\mathbf{w}^j)\right\|^2\right]\leq \frac{F(\mathbf{w}^J)-F^*}{c\alpha\bar{U}J}+\frac{1}{c}\left[\frac{\alpha U_{max}L}{2KB\bar{U}}\sigma_L^2 + \frac{5\alpha^2U_{max}^3L^2}{2\bar{U}^2}(\sigma_L^2+6U_{max}\sigma_G^2)\right]\label{theo1_proof}.
\end{align}
The proof is then complete.

\end{document}